\newcommand{\sota}{{\textit{state-of-the-art}}}

\newcommand{\Ours}{{\textit{MV-DeepSimNets}}}
\newcommand{\OursBase}{{\textit{DeepSimNets}}}
\newcommand{\Unet}{U-Net }

\documentclass[10pt,letterpaper]{article}

\usepackage{cvpr}
 
\definecolor{cvprblue}{rgb}{0.21,0.49,0.74}
\usepackage[pagebackref,breaklinks,colorlinks,allcolors=cvprblue]{hyperref}

\usepackage{amssymb}
\usepackage{array}
\usepackage{amsthm}
\usepackage{xcolor}
\usepackage{float}
\usepackage{graphicx}
\usepackage{amsmath}
\usepackage{comment}
\usepackage{ragged2e}
\usepackage{soul}
\usepackage{caption}
\usepackage{subcaption}
\usepackage{adjustbox}
\usepackage{makecell}
\usepackage{multirow}
\usepackage{tikz}
\usepackage{graphicx}

\usepackage{lineno}
\newcommand*{\bigL}{\scalebox{1.5}{\ensuremath{\mathcal{L}}}}

\usepackage[figuresright]{rotating}

\usepackage{xcolor}
 
\begin{document}
\title{\vspace{1.0cm}Multi-view dense image matching with similarity learning and geometry priors}


\author{Mohamed Ali Chebbi\textsuperscript{1}
\and
Ewelina Rupnik\textsuperscript{1}
\and
Marc Pierrot-Deseilligny\textsuperscript{1} $\quad$ Paul Lopes\textsuperscript{2}
\and
{
 	\textsuperscript{1}Univ Gustave Eiffel, LASTIG, ENSG-IGN, F-94160 Saint-Mandé, France \hspace{0.5cm}
	\textsuperscript{2}Thales, France 
}
\vspace{1cm}
}
\providecommand{\keywords}[1]{\textbf{\textit{Keywords--}} #1}
\vspace{1cm}
\maketitle
\begin{abstract}
We introduce \Ours, a comprehensive suite of deep neural networks designed for multi-view similarity learning, leveraging epipolar geometry for training. Our approach incorporates an online geometry prior to characterize pixel relationships, either along the epipolar line or through homography rectification. This enables the generation of geometry-aware features from native images, which are then projected across candidate depth hypotheses using plane sweeping. Our method's geometric preconditioning effectively adapts epipolar-based features for enhanced multi-view reconstruction, without requiring the laborious multi-view training dataset creation. By aggregating learned similarities, we construct and regularize the cost volume, leading to improved multi-view surface reconstruction over traditional dense matching approaches. \Ours~demonstrates superior performance against leading similarity learning networks and end-to-end regression models, especially in terms of generalization capabilities across both aerial and satellite imagery with varied ground sampling distances. 
Our pipeline is integrated into  MicMac software and can be readily adopted in standard multi-resolution image matching pipelines. 
\end{abstract}



\vspace{2cm}


\section{Introduction}
\label{INTRO}
\sloppy
{Along with the improving performance of stereo matching algorithms for 3D surface reconstruction, the number of applications where scene's geometry is of interest is also growing. Three-dimensional geometries are indispensable in 3D scene understanding for autonomous navigation, large scale city mapping and digital twins, infrastructure monitoring, as well as in environment-oriented case studies such as precision agriculture or forest canopy growth estimation. This tendency further drives the algorithms to provide higher quality surfaces in terms of correctness, completeness, and robustness against various geometric and radiometric acquisition conditions. Traditional stereo matching based on handcrafted features and surface smoothness constraints \cite{micmac,Hirsh} performs well as long as the scene has a discriminative texture, is Lambertian and free of occlusions. Modern convolutional neural networks (CNNs) have more expressive capacity as they replace the heuristics encoded within handcrafted features with arbitrary nonlinear functions that are learned from data itself. This, combined with large receptive fields, makes CNN-based stereo matching strategies \cite{zbontar2016,konrad2017,Han2015,Zagoruyko2015} often better performing, in particular in the challenging scenarios mentioned above.
}

{Consequently, many learning-based approaches have been developed in recent years. One can distinguish two categories of approaches: hybrid and end-to-end; stereo and multi-view stereo. Hybrid methods follow the traditional two-step matching pipeline except the handcrafted features are replaced with learnt features (i.e., learnt image representations), or learnt similarities. The end-to-end methods regress depths directly from image intensities. Architectures that regress depths end-to-end have more powerful learning capacity and are capable of implicitly learning scene's likely shapes. However, the 3D CNNs employed within such architectures entail holding in memory the cost volume representing the entire scene, hence are extremely greedy with memory. This impedes their use in large scale 3D reconstruction projects (e.g., depths from high resolution satellite images). Moreover, unlike hybrid methods trained exclusively from image intensities and, therefore, by nature more generic and not tied to a particular landscape, end-to-end networks are bound by the size and typology of the scene it was trained on \cite{WU2024103715}.
}

{Engaging multiple views in stereo matching is known to be advantageous because it adds observation redundancy contributing to lower depth noise and more complete surfaces. However, choosing a multi-view setting requires dealing with camera poses (i.e., extrinsic and intrinsic parameters) and searching for correspondences along paths which are nonparallel to the image coordinate frame. This makes training, inference and fabricating multi-view ground truth datasets substantially more laborious compared to the stereo case using epipolar images.}
{Moreover, despite the remarkable performance of modern regression-based stereo matching -- particularly in enhancing disparity predictions along object boundaries~\cite{PSMNet} and recovering high-frequency details~\cite{raftstereo} -- these networks remain inherently limited to the epipolar case, making a direct multi-view extension infeasible. The only way to incorporate regressed stereo disparities into a multi-view framework is through post-processing, which involves selecting epipolar pairs, estimating disparities, warping them to a common coordinate system, and subsequently fusing the pairwise results. In contrast, hybrid approaches learn representations that are less constrained by geometric assumptions, enabling direct multi-view cost volume construction by leveraging prior knowledge of camera poses.}

Our objective is depth prediction for digital surface model (DSM) computation from aerial and satellite imagery. We focus on stereo matching approaches that are (i) generic and insensitive to image acquisition geometry, (ii) easily transferable to new scenes without the need for exhaustive re-training; and (iii) handle multi-view configurations. To address those requirements, we propose a framework that relies on representation (i.e., embedding) learning and incorporates multi-views, however, only at inference time. By circumventing the multi-view training phase we can capitalize on the abundant \textit{state-of-the-art} stereo benchmarks and reduce the complexity of the entire pipeline. At inference,  multiple images are integrated thanks to available geometric \textit{a priori} such as the camera poses. 
{More precisely, we transition between the native image geometry and the epipolar geometry using known transformations. Features are extracted from the images after they have been transformed to the epipolar geometry, and are subsequently mapped back to the original image geometry via the inverse transformation. Surface inference is performed by sweeping a plane through the scene at predefined depth hypotheses, which are backprojected to compute pixelwise multiview similarities using the feature maps.}
This work is an extension of \OursBase~\cite{Chebbi_2023_CVPR} and we refer to it throughout the paper as multi-view \OursBase~or \Ours.

Our main contributions are as follows:
\begin{itemize}
    \item We propose a multi-view dense image matching framework specifically designed for aerial and satellite imagery, leveraging deep features trained on epipolar image pairs.
    \item  Unlike existing multi-view learning-based approaches, our framework does not require additional re-training, making it readily deployable.
    \item We demonstrate that representations learned from epipolar pairs are highly generalizable, effectively transferring to unseen landscapes and enabling large-scale deployment.
\end{itemize}

\section{Related Work}

\paragraph{Multi-view dense image matching} End-to-end learnable approaches can be categorized into volumetric and depth map reconstructions. The former warp all RGB overlapping images into a plane-sweep volume, followed by feature extraction, patch matching and cost aggregation \cite{ji2017surfacenet,huang2018deepmvs}. Such volumetric representation is extremely demanding in terms of memory requirements and therefore handle only small-scale reconstructions. More recent depth map based multi-view approaches follow the pipeline of (1) feature extraction, (2) feature to volume warping, (3) feature volume to feature cost aggregation, (4) regularisation and (5) depth regression. The two standard feature backbones are Feature Pyramid Network \cite{lin2017feature} or transformer-based (i.e., ViT) representations  \cite{dosovitskiy2010image,cao2022mvsformer,liu2023epipolar,ding2022transmvsnet,weinzaepfel2023croco}. The latter are said to perform better on textureless areas but require careful training and large training datasets. 
The features are classically warped at constant depths with a differential homography \cite{mvsnet} or RPCs \cite{rpcwarping}  over a volume spanning a portion of the scene's 3D volume visible to the so-called reference image. Next, the pair-wise similarities can be computed using a variance-based metric \cite{mvsnet} or the group-wise correlation \cite{xu2020groupcorr}, while incorporating multiple views comes down to averaging over multiple cost volumes. Additionally, a per-view and per-pixel weighted average can help infuse visibility information \cite{wang2021patchmatchnet}. Applying 3D UNet-like CNNs \cite{mvsnet}, recurrent smoothing with convolution GRU \cite{rmvsnet2019} or boundary-aware spatial regularization \cite{wang2021patchmatchnet} to the aggregated cost volume regularises the costs thus suppresses noise. Finally, the depths are regressed through a \textit{softmax} operation \cite{kendall2017end} or calculated as a combination of regression (\textit{softmax}) and classification (\textit{argmax}), known as temperature-based depth prediction \cite{cao2022mvsformer}. An ultimate depth refinement step can take the form of post-process filtering \cite{mvsnet}, a depth residual network extended with an RGB image \cite{wang2021patchmatchnet}, entropy-based confidence map \cite{ma2021epp}, crafted 3D point consistency loss \cite{kim2021just}. 

\paragraph{Geometry priors} Multi-view image matching relaxes some of the underlying geometry priors compared to epipolar geometry. Feature extraction is performed on images in their native geometry, where across-view correspondences undergo displacements in 2-dimensional space, as opposed to 1-dimensional displacements as in the epipolar case. This introduces additional strain on the learning task, as the network must learn representations invariant to those transformations and be able to identify more complex matching relationships. To achieve good network robustness, many views participate in the training \cite{wang2021patchmatchnet}, setting more stringent requirements on the train dataset and inevitably increasing train times. This can be considered an impediment in satellite or airborne acquisitions where the number of views is limited, and the images are of very high resolution (e.g., Pléiades 1AB swath at 40k pixels). Because the latter represents our potential input data, in \Ours~we follow the training procedure based on epipolar images as presented in \cite{Chebbi_2023_CVPR} and compute multi-view similarity scores outside the learning architecture by on-the-fly warping of the native images to a common geometry, be it an epipolar geometry or a common plane.

\paragraph{Towards memory efficiency} Building and regularizing cost volume for large disparity ranges and high resolution images is the bottleneck of end-to-end multi-view matching algorithms. Consequently, there have been many attempts to curb the memory footprint. In \cite{mvsnet} both the images and the cost volume are downscaled at the expense of downgraded depth precision; \cite{rmvsnet2019} sequentially regularises many 2D cost maps with recurrent convolutional GRU, while \cite{gu2020cascade} was among the first to embed depth prediction within a learning multi-resolution framework similar to that of the classical MVS pipelines and referred to as the cascade cost volume. Similar coarse-to-fine strategies were adopted by, e.g., \cite{ding2022transmvsnet,kim2021just}. Differently, \cite{liu2023epipolar} reduces the computational overhead of ViTs which attend to all pixels in the image by constraining it to corresponding lines, i.e. Epipolar Transformer, while \cite{ma2021epp} exploits the epipolar geometry to reduce the cost volume footprint and avoid spatial resolution downsampling. \Ours~is built in a classical multi-resolution pipeline \cite{micmac} where depths are initialized at low resolution with cross-correlation and SGM, and our learnt representations intervene at later stages. Hence, our network is deployed over a cost volume representing a narrow envelope around the initialized depths.

\paragraph{Supervised and unsupervised approaches} Driven by the complexity of providing a high-quality ground truth depth, some works focus on unsupervised training frameworks. Many authors went in the footsteps of the optical flow self-supervised approaches \cite{yu2016back} to estimate stereo disparity \cite{zhou2017unsupervised}. A typical end-to-end self-supervised network consists of a feature extraction module, a correlation layer, cost aggregation / filtering and disparity extraction. Starting from random initializations or from predictions given by a pre-trained net \cite{weinzaepfel2013deepflow}, self-supervision minimizes a photometric loss, oftentimes accompanied by a smoothness or a semantic loss \cite{yang2018segstereo}. Video acquisitions with small base-to-height ratios or stereo-camera rigs are suitable for providing training datasets for self-supervision. Nevertheless, repetitive patterns and homogeneous regions remain challenging for fully unsupervised methods \cite{unsupervisedstereo}. Alternatively, semi-supervision with a few sparse points not only brings down the burden of providing ground truth disparities, but also improves generalization to new scenarios \cite{kim2021just}. 
Recently, neural radiance fields (NeRF) have shown decent capabilities to produce convincing new views and depth maps by exclusively relying on a photometric loss \cite{mildenhall2021nerf}. However, as of now reconstructing a NeRF representation from a few viewpoints is far from outperforming traditional stereo matching approaches. A promising research direction is to exploit the complementarity of the two methods, stereo matching for robustness with few views and NeRF-like representation for better handling of occlusions and non-Lambertian surfaces \cite{chang2022rc,prinzler2023diner}. \Ours~learns feature representation in a self-supervised manner where the ground truth depths obtained with LiDAR provide training samples but are not considered in the network's architecture or the loss.

\begin{figure}[tbh]
  \centering
   \includegraphics[width=0.8\linewidth]{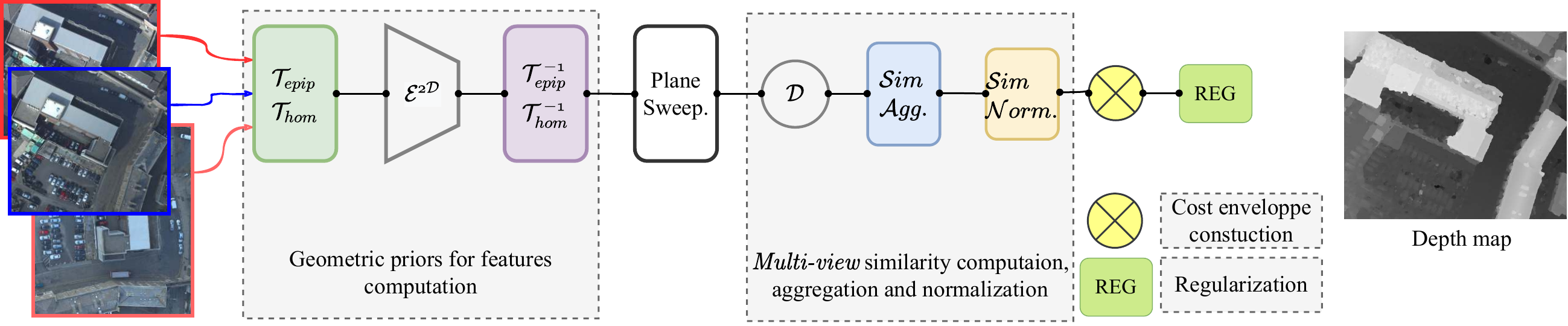}
    \caption{\textbf{Mutli-view multi-resolution neural image matching pipeline}. The reference image is outlined in \textcolor{blue}{$\square$}, and the query images are in \textcolor{red}{$\square$}. First, pairs of images are transformed to rotation-aligned positions either with epipolar $\mathcal{T}_{epip}$ or with homographic $\mathcal{T}_{hom}$ warping and fed to the feature extractor ($\mathcal{E^{\text{2}D}}$). The resulting epipolar feature maps are warped back by applying $\mathcal{T}^{-1}_{epip}$ or $\mathcal{T}^{-1}_{hom}$, followed by the plane sweeping to compute features at different depth hypotheses $\mathcal{Z}$.
    Stereo-pair similarity at a depth hypothesis is computed with cosine similarity or by feeding concatenated feature maps to the decision MLP ($\mathcal{D}$). The multi-view similarity score at depth $\mathcal{Z}$ is averaged over all stereo-pairs observing that scene. Finally, the optimal depths are found with SGM.} 
  \label{fig:mutlimulti}
\end{figure}

\section{Method}
\label{METHOD}
\sloppy

\begin{figure}[htbp]
    \centering
        \includegraphics[width=0.9\linewidth]{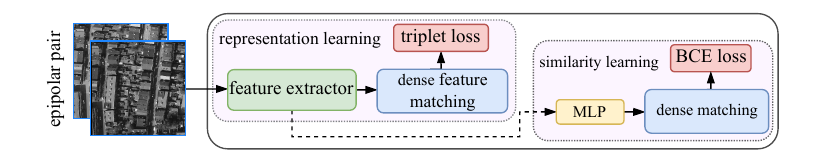}
    \caption{\textbf{Epipolar Similarity Learning Pipeline}. Matching is carried out on learned representations by measuring the cosine distance between features. Alternatively, the MLP infers learned similarities.}
    \label{fig:deepsimnet_pipe}
\end{figure}

{In the following, we present the methodology behind \Ours~(see \Cref{fig:mutlimulti}). We begin by introducing \OursBase, which serve as the baseline feature backbone within our multiview framework. Next, we describe two approaches transforming epipolar-based features into a multiview setting (see \Cref{fig:epipwarpingrequirements}). We also outline the plane-sweeping process used in feature retrieval during inference (see \Cref{fig:placesweep}). 
}
{
\subsection{Preliminaries on epipolar similarity learning}\label{subsec:prelimin} The training scheme of \OursBase~comprises two main steps: representation learning and similarity learning, as illustrated in \Cref{fig:deepsimnet_pipe}. First, pairs of large epipolar images are processed through a CNN-based feature backbone. A sample mining procedure then selects positive and negative sample pairs based on feature maps and ground truth (GT) disparities. Hence, we self-supervise the network as the GT disparities serve exclusively to provide matching and non-matching samples. Finally, a triplet (representation learning) or binary cross entropy (similarity learning) loss is applied, encouraging matching samples to be close while pushing non-matching samples apart in the feature space.
}

{\paragraph{Feature backbones} We trained 3 variants of feature extractors: a U-Net32, an attention U-Net~\cite{att_unet} and MS-AFF. The latter is an explicit mutli-scale self-attentional feature learning and fusion network that is 8$\times$ and 10$\times$ lighter than U-Net32 and U-Net attention variants, respectively. MS-AFF leverages robust multi-scale features by means of an iterative attentive feature fusion mechanism. For further details, please refer to the full paper \cite{Chebbi_2023_CVPR}.} 

{\paragraph{Iterative negative feature sampling} While our sampling approach employs feature triplets for contrastive learning, ensembling feature sets simultaneously for a large-size epipolar image naturally leads to the well-known \textit{n-pair} loss~\cite{NIPS2016_6b180037}. The n-pair loss compares a \textit{matching} feature pair against a set of \textit{negative} features, including other \textit{positives} and by doing so it enhances feature distinctiveness and promotes the \textit{one-to-one} mappings of feature positives. Note that the \textit{n-pair} comparison is not explicitly enforced in our framework. Instead, by dynamically updating negative examples throughout training, our matching features are continuously challenged against nearly all negatives within the predefined neighborhood. Take the set features $f_{1}$,$f_{2}$,$f_{3}$ and their matching features $f^{+}_{1}$,$f^{+}_{2}$,$f^{+}_{3}$ in the left epipolar line in \Cref{fig:convergencetonpairloss}. For a pair $\{ f_{2},f^{+}_{2}\}$, we consider the $\left[ 1,4 \right]$ \textit{negative} pixel sampling interval. Notice that after 7 iterations, only $f_{-2}$ and $f_{6}$ were omitted. The remaining of the neighboring features along the epipolar line, including $f^{+}_{1}$ and $f^{+}_{3}$, appear as \textit{mismatches} and are accounted for in our loss function. {Since all our models training scenarios are run for 50 epochs (i.e iterations)~\cite{Chebbi_2023_CVPR}, the proposed sampling naturally spans almost all neighboring non-matching pixels and convergence towards the n-pair loss is enforced.}}

\definecolor{blueblue}{rgb}{0.85, 0.9, 0.99}
\definecolor{redred}{rgb}{0.97, 0.8, 0.8}
\definecolor{greengreen}{rgb}{0.83,0.9,0.83}
\definecolor{purplepurple}{rgb}{0.88,0.83,0.9}
\begin{figure}[htbp]
    \centering
    \includegraphics[width=0.7\textwidth]{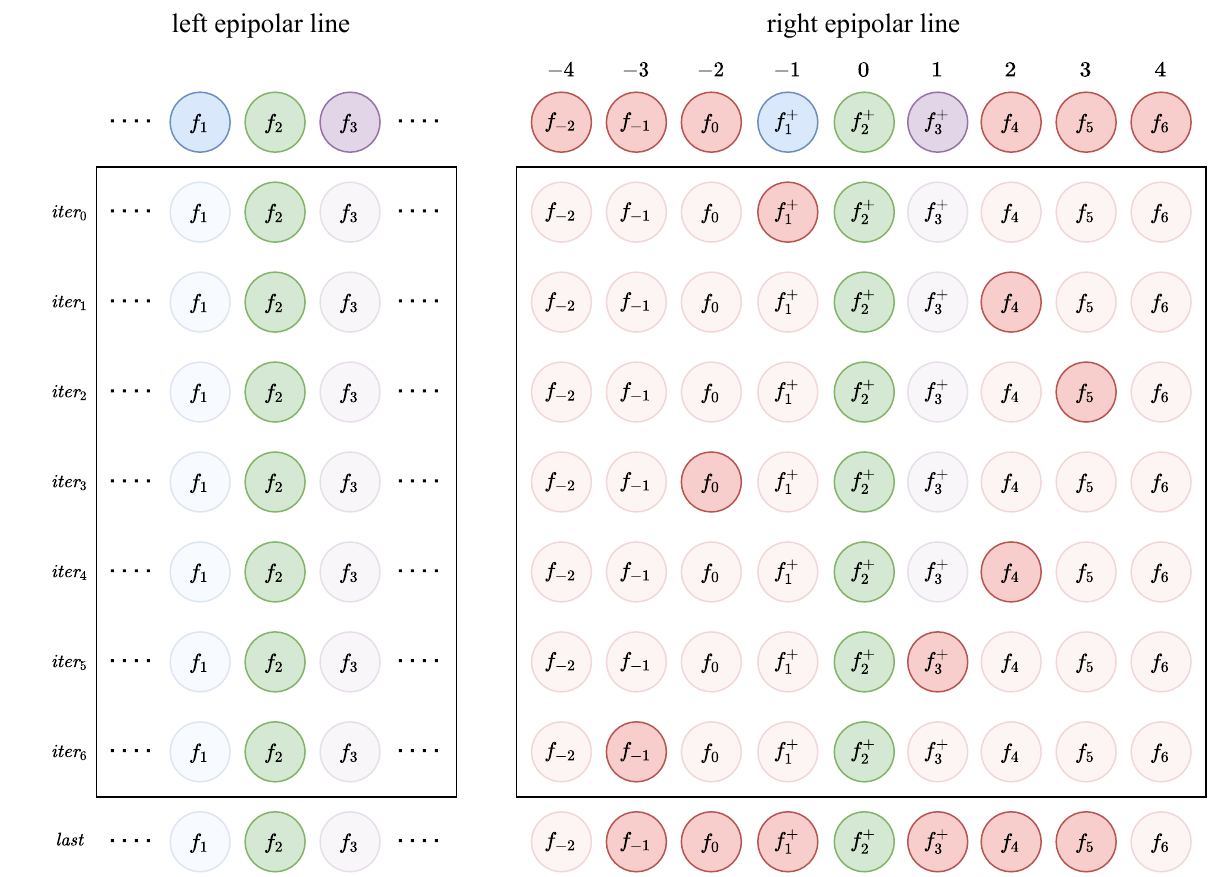}
    \caption[Iterative Negative Sampling.]{\textbf{Iterative Negative Sampling.} \colorbox{blueblue}{\{$f_{1}$,$f^{+}_{1}$\}}, \colorbox{greengreen}{\{$f_{2}$,$f^{+}_{2}$\}}, \colorbox{purplepurple}{\{$f_{3}$,$f^{+}_{3}$\}} are feature matches. By focusing on the \colorbox{greengreen}{\{$f_{2}$,$f^{+}_{2}$\}} pair, iterating on negative samples leads to \colorbox{greengreen}{$f_{2}$} being compared to almost all \colorbox{greengreen}{$f^{+}_{2}$} \colorbox{redred}{neighboring features}. We can assume that by iteratively sampling negatives, Our loss function converges to the \textit{n-pair} loss that involves multiple negatives instead of one.}
    \label{fig:convergencetonpairloss}
\end{figure}

{\paragraph{Losses} We use a triplet loss to learn feature representations and a binary cross-entropy loss to train the similarity decision. To prevent high similarity scores in occluded areas, we regularize the loss in those areas with an additional penalty term (see \Cref{eq:triocc,eq:bceocc}). During training, we follow a differential training scheme: we first train the feature representations independently, then freeze them while training the similarity function. Our triplet $\bigL_{3_{All}}$ and BCE $\bigL_{BCE_{All}}$ losses are as follows:
}

\begin{equation}
\label{eq:triocc}
\bigL_{3_{All}} =\underbrace{\hspace{-0.1cm}\sum_{(i,j) \in \mathcal{X}_{nocc}}\hspace{-0.2cm}\max(\mathcal{S}^{i,j}_{-}- \mathcal{S}^{i,j}_{+}+ m ,0)\hspace{0.1cm}}_{\textit{non occlusion term}}+\hspace{-0.1cm}\underbrace{\sum_{(i,j) \in \mathcal{X}_{occ}}\hspace{-0.2cm}\max(\mathcal{S}^{i,j}_{1-}+\mathcal{S}^{i,j}_{2-},0)}_{\textit{occlusion term}}~,
\end{equation}
where $\mathcal{S}^{i,j}_{+}=\left<f^{i,j}_{nocc},f^{i,j}_{+}\right>$ and $\mathcal{S}^{i,j}_{-}=\left<f^{i,j}_{nocc},f^{i,j}_{-}\right>$ are cosine similarities between a reference~($_{R}$) not-occluded descriptor $f^{i,j}_{nocc}$ and a matching $f^{i,j}_{+}$ and non matching $f^{i,j}_{-}$ query~($_{Q}$) descriptors. $\mathcal{S}^{i,j}_{1-}=\left<f^{i,j}_{occ},f^{i,j}_{1-}\right>$ and $\mathcal{S}^{i,j}_{2-}=\left<f^{i,j}_{occ},f^{i,j}_{2-}\right>$ are cosine similarities between a reference occluded descriptor $f^{i,j}_{occ}$ and two query nonmatching descriptors $f^{i,j}_{1-}$ and $f^{i,j}_{2-}$. The margin m was empirically fixed to {0.3} for all experiments. The MLP ($\mathcal{D}$) concatenates a reference and a query descriptors and outputs a learned similarity score. Similarly, an occlusion-aware binary cross entropy loss~(see Equation~\ref{eq:bceocc}) is used to train the MLP. 
\begin{equation}
\label{eq:bceocc}
\bigL_{BCE_{All}}=\underbrace{-\hspace{-0.3cm}\sum_{(i,j) \in \mathcal{X}_{nocc}}\hspace{-0.25cm}\mathcal{Y}^{i,j}_{-} \log(1-\mathcal{S}^{i,j}_{-}) + \mathcal{Y}^{i,j}_{+} \log(\mathcal{S}^{i,j}_{+})}_{\textit{non occlusion term}}\hspace{0.1cm}\underbrace{-\hspace{-0.3cm}\sum_{(i,j) \in \mathcal{X}_{occ}}\hspace{-0.25cm}\mathcal{Y}^{i,j}_{-} \Bigl(\log(1-\mathcal{S}^{i,j}_{1-}) + \log(1-\mathcal{S}^{i,j}_{2-})\Bigr)}_{\textit{occlusion term}}~,
\end{equation}
here, $\mathcal{S}^{i,j}_{+}=\mathcal{D}(f^{i,j}_{nocc},f^{i,j}_{+})$, $\mathcal{S}^{i,j}_{-}=\mathcal{D}(f^{i,j}_{nocc},f^{i,j}_{-})$, $\mathcal{S}^{i,j}_{1-}=\mathcal{D}(f^{i,j}_{occ},f^{i,j}_{1-})$ and $\mathcal{S}^{i,j}_{1-}=\mathcal{D}(f^{i,j}_{occ},f^{i,j}_{1-})$ are learnt similarity scores. $\mathcal{Y}_{+}$ and $\mathcal{Y}_{-}$ are binary masks that describe matching and non-matching pixel locations. \newline

\subsection{Multi-view image matching}
Our multi-view multi-resolution neural image matching pipeline, referred to as \Ours, is depicted in Figure~\ref{fig:mutlimulti}. It takes $N$ views along with their corresponding camera poses as input and produces a dense depth map as output. As a hybrid approach, the main computational components of \Ours~include geometry-aware feature extraction, similarity computation, and regularization \cite{Hirsh}. Our work focuses on features and similarities, proposing that features derived from stereo configurations can be adapted for multiview scenarios, thereby circumventing the need for tedious network retraining using multiple views. To this end, we employ \OursBase~\cite{Chebbi_2023_CVPR} introduced above.  
Extending this framework to a multiview context entails integrating a transformation, or a \textit{ geometric prior} as we refer to it hereafter, that maps the native image geometry to a representation free of rotation ambiguity. We accomplish this through epipolar $\mathcal{T}_{epip}$ or homographic $\mathcal{T}_{hom}$ warping. In contrast to \sota~multiview learning-based approaches \cite{huang2018deepmvs,cao2022mvsformer,rpcwarping}, we situate geometric operators external to the learning module. By eliminating the rotation variance, we explicitly provide directional information of image rays to the pipeline and remove the necessity to learn rotation-invariant features from the data. Consequently, smaller networks can be used for feature learning purposes.

\begin{figure}[t!]
    \centering
    \includegraphics[width=0.75\linewidth]{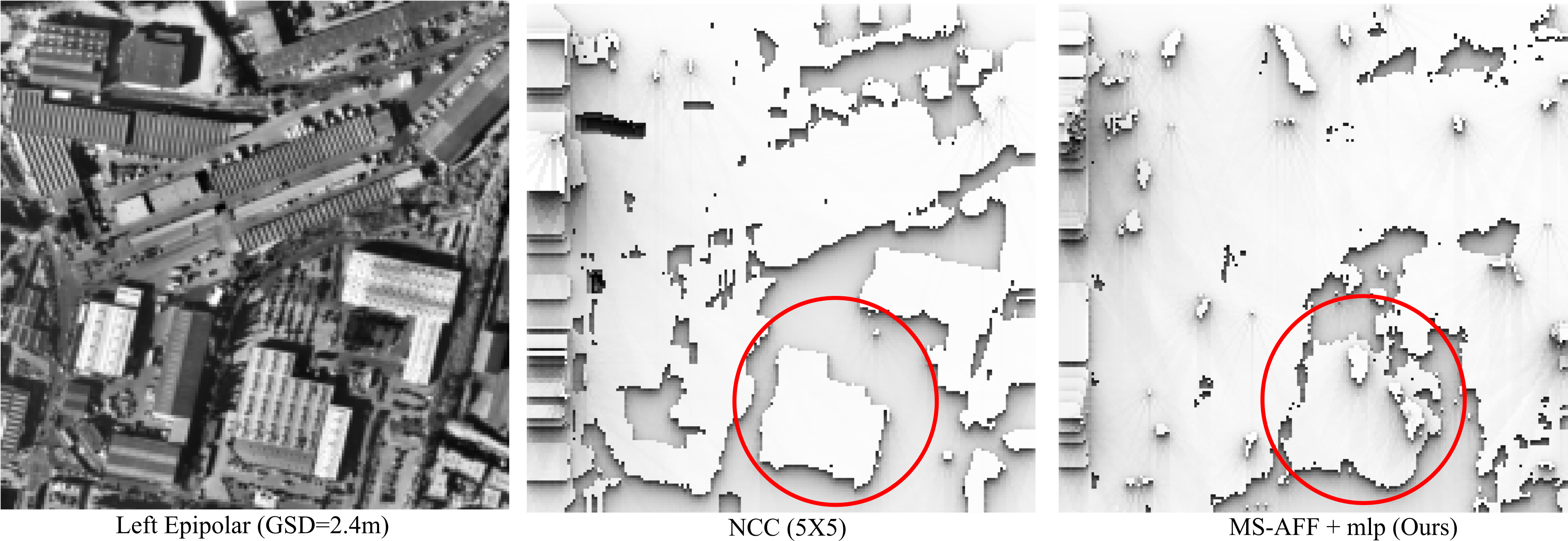}
    \caption{\textbf{Normalized Cross Correlation (NCC) vs MS-AFF at Low Resolutions.} Enabling NCC at lower resolutions produces accurate surface predictions (\textcolor{red}{$\bigcirc$}). In contrast, MS-AFF, not having been trained on lower GSD imagery, struggles to generate reliable reconstructions at 2.4m GSD.}
    \label{fig:explain_why_ncc_good_low_resolution}
\end{figure}

{Similar to \OursBase, we adopt an iterative multi resolution depth prediction pipeline, where low-resolution depth maps are computed using handcrafted features (e.g., NCC), while learned features are introduced at higher resolutions. This approach is motivated by two key factors. First, low-resolution images contain limited information and exhibit low expressivity, reducing the robustness of the learned features. In such cases, handcrafted features are more reliable, as illustrated in Figure~\ref{fig:explain_why_ncc_good_low_resolution}. Using this observation, our pipeline remains frugal, requiring fewer training datasets and reducing training times.
Second, \OursBase~was trained on aerial images with small pixels (GSD $\sim4$ cm). Given the multiscale nature of our features, the model has encountered image patches up to four times this size ($\sim16$ cm). However, during hierarchical depth prediction on satellite images, the resolution gap becomes substantial. For instance, in a WorldView-3 acquisition at 30 cm GSD, our pipeline downscales images by a factor of 8, reaching an effective resolution of 2.4 m—well beyond what the model has seen during training. Unsurprisingly, without further training, performance degrades. Even if retraining were an option, existing satellite datasets suffer from LiDAR registration accuracies as poor as 1.8 pixels \cite{patil}, making them unsuitable for use in training.}

\begin{figure}[tb]
    \centering
    \subfloat[][Epipolar warping]
    {
    \includegraphics[width=0.49\columnwidth]{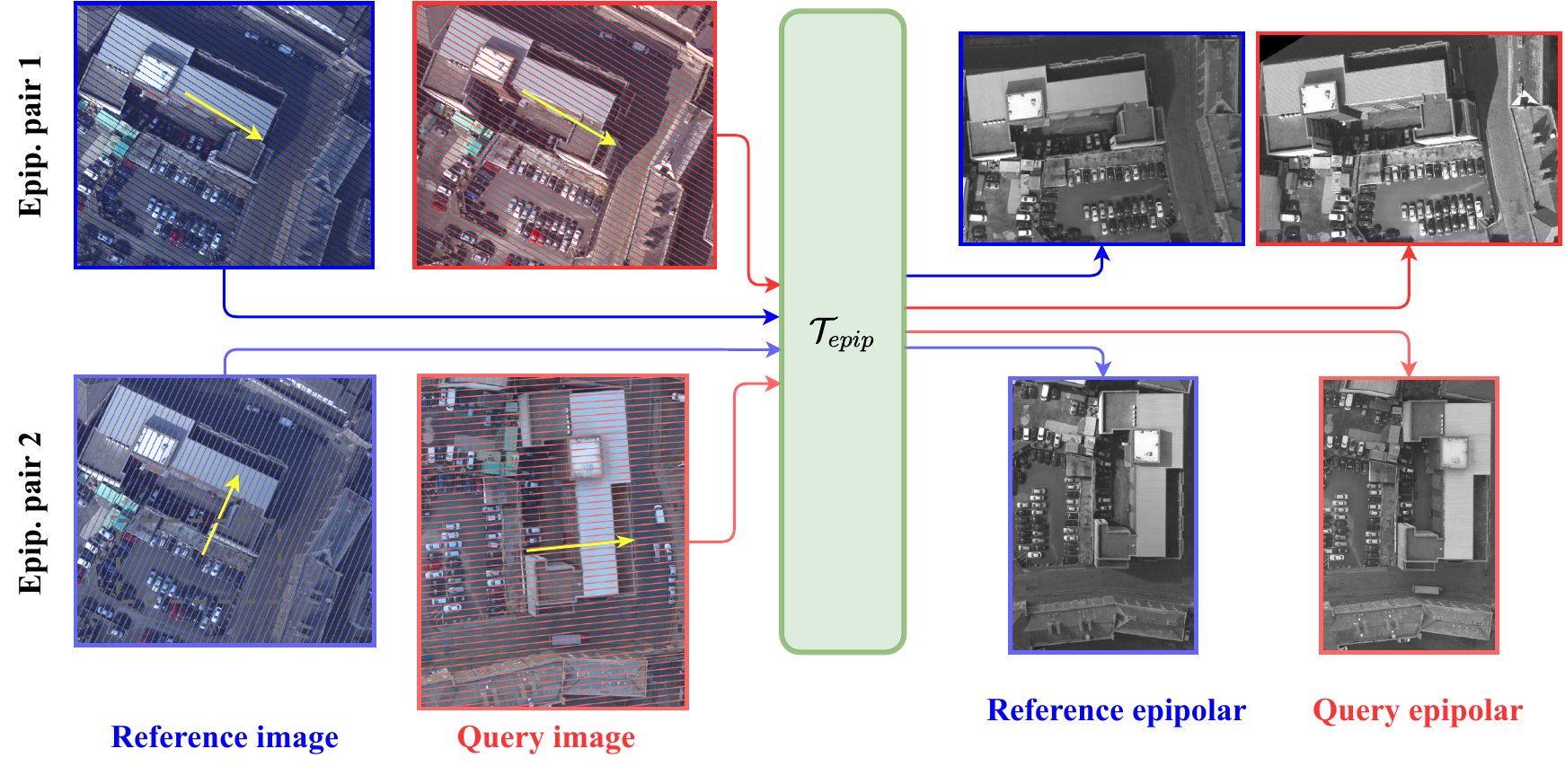} \label{fig:epipolarwarping} 
    }
    \subfloat[][Homography warping]
    {
    \includegraphics[width=0.49\columnwidth]{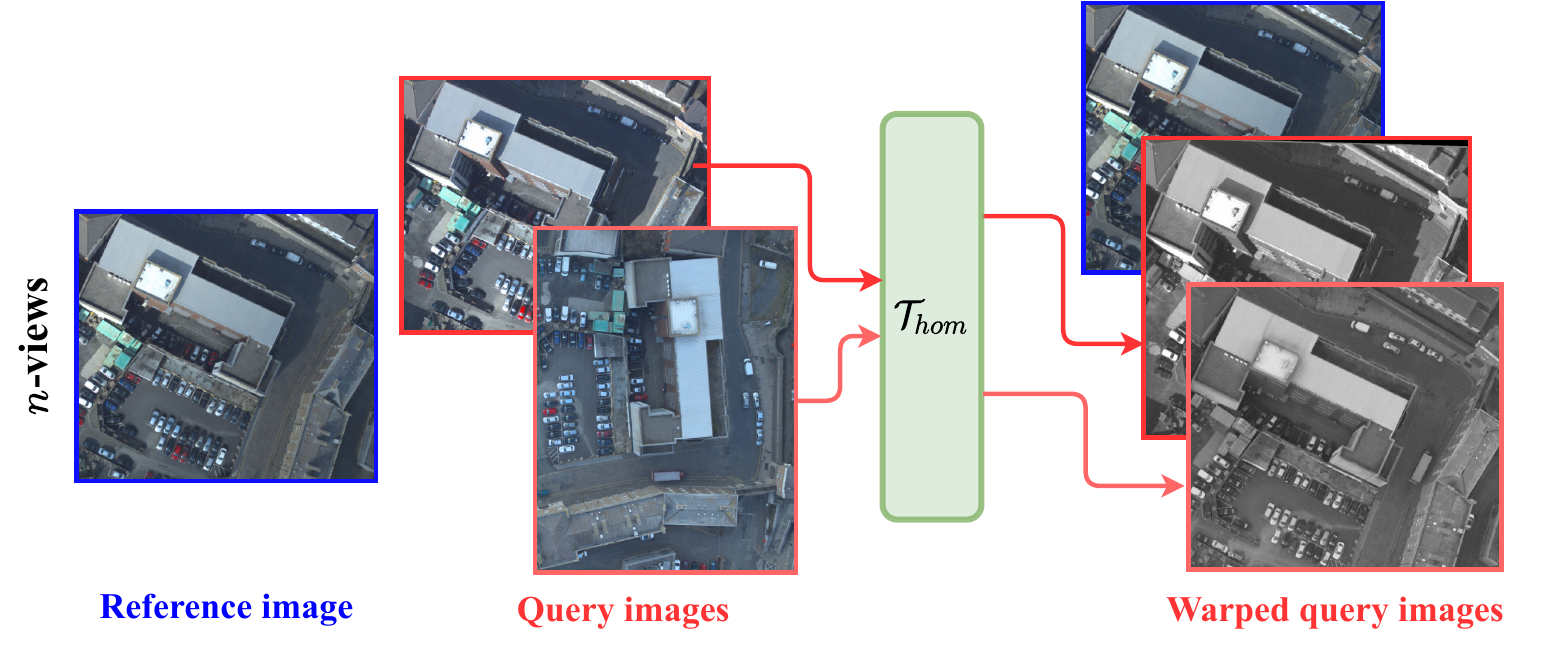} \label{fig:homographicwarping} 
    } \qquad
    \caption{\textbf{Geometry-aware features creation.} To leverage feature extractors trained on epipolar images in a multi-view context, we transform our input images to an equivalent geometry that mimics the epipolar configuration. We achieve this through either epipolar $\mathcal{T}_{epip}$ or homography $\mathcal{T}_{hom}$ warping. Feature computed in the equivalent geometry are then warped back to their respective native geometries and used as such in the subsequent similarity scoring. Note that the homography can handle a general multi-view configuration, as opposed to multi-view stereo possible with the epipolar warping.}
    \label{fig:epipwarpingrequirements}
\end{figure}

\begin{figure}[tb]
  \centering     
     \subfloat[][A plane hypothesis and its corresponding 2D rectification grid.]{\includegraphics[height=4.17cm,width=7.4cm]{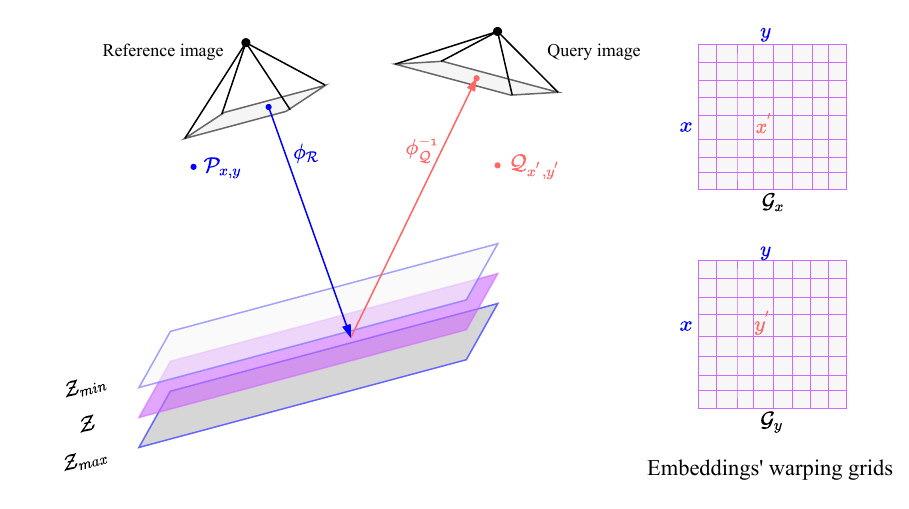}\label{fig:planesweep}}
     \hspace{1cm}
     \subfloat[][Features' visualization with and without the geometry prior (\textit{epip}).]{\includegraphics[height=4.17cm,width=7cm]{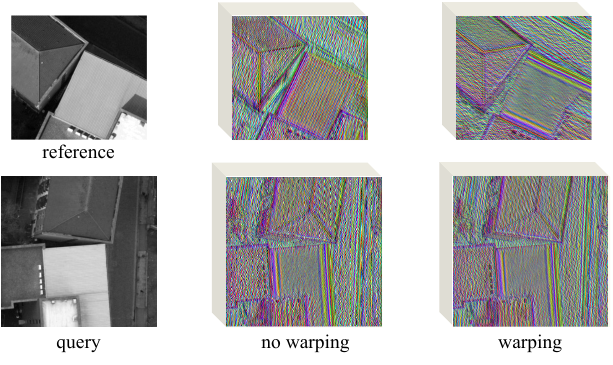}\label{fig:epipolarfeaturesaware}}
    \caption{\textbf{Plane sweeping.} At inference time query features are rectified to the reference image with the help of scene's plane hypotheses.}\label{fig:placesweep}
\end{figure}

\paragraph{Geometry-aware features}
In multiview acquisitions, images can take arbitrary positions and rotations. Without careful training across a range of image rotations, CNN feature extractors struggle to compute distinctive image representations \cite{transferabilityoffeatures}. An experiment in Figure~\ref{fig:epipolarfeaturesaware} underscores the rotation sensitivity of a CNN feature extractor trained on epipolar pairs, where features extracted from two images with arbitrary rotation shift exhibit inconsistencies. It suggests that CNNs implicitly capture epipolar cues during training. We exploit the fact that CNNs are invariant to translation and, when trained on epipolar images, to a certain degree to perspective deformations. Precisely, our guidance relies on transforming original images to geometries where those invariance properties are preserved.
{To do that, we adopt two geometric transformations: epipolar geometry $\mathcal{T}_{epip}$ and homography $\mathcal{T}_{hom}$ (see Figure~\ref{fig:epipwarpingrequirements}). Epipolar geometry transforms a stereo-pair to a position where corresponding pixels run along the rows. Depending on the camera model, $\mathcal{T}_{epip}$ is computed with either a calibrated epipolar rectification (i.e., pinhole) \cite{fusiello2000compact} or a generic algorithm based on point correspondences (e.g., pushbroom) \cite{epipolarresampling} . The homography transforms the query images to the geometry of the reference image. It is a plane to plane transformation which implies that the rectified images superpose at some mean depth of the scene. Anything outside that depth incurs disparities across the overlapping images, however, for aerial and satellite acquisitions, such disparities remain small. We calculate the 8-parameter homography $\mathcal{T}_{hom}$  using sparse point correspondences (e.g., SIFT \cite{lowe1999object}) and a robust estimator. Note that  unlike the \sota~ end-to-end methods for image matching \cite{rpcwarping,rmvsnet2019,huang2018deepmvs,cao2022mvsformer}, we do not use homography for plane hypothesis testing.}
%
{The features are calculated following the workflows illustrated in Figure~\ref{fig:epipwarpingrequirements}. An image in its native geometry~$\mathcal{I}$ is first transformed to an \textit{aligned} geometry with either epipolar or homography transformation~$\mathcal{T}$: }

\begin{equation}
\label{eq:epipolarresampling}
\begin{array}{ll}
\mathcal{I_{T}}=\mathcal{T}\left(\mathcal{I}\right).
\end{array}
\end{equation}
Oriented features are then computed with \OursBase, followed by warping back to the original image space with the inverse operators:
\begin{equation}
\label{eq:epipolarwarping}
\begin{array}{ll}
\mathcal{F}_{I_{R}}=\mathcal{T}^{-1}_{R}\Bigl[\mathcal{E}^{2D}\Bigl(\mathcal{I_{T_R}}\Bigr)\Bigr], \\
\mathcal{F}_{I_{Q}}=\mathcal{T}^{-1}_{Q}\Bigl[\mathcal{E}^{2D}\Bigl(\mathcal{I_{T_Q}}\Bigr)\Bigr], 
\end{array}
\end{equation}
where  $\mathcal{F}_{I_{R}}$ and $\mathcal{F}_{I_{Q}}$ are reference and query feature maps, respectively; $\mathcal{E}^{2D}$ is the feature extractor; $\mathcal{T}^{-1}$ is the inverse rectification operator. Note that we assume the operators to be bijective and continuous, thus we can compute their inverses. At inference, the plane sweeping is operated on the transformed features which endow the geometric priors.
\paragraph{Multi-view plane sweeping}
Our image matching is performed in the geometry of the reference image. For every pixel position in that image, we follow the standard plane sweeping approach to explore the 3D scene \cite{collins1996planesweep} (see Figure~\ref{fig:planesweep}). With known camera poses $\phi$, at each hypothesis $\mathcal{Z}$, we ortho-rectify the features $\mathcal{F}_{I_{Q}}$ of the query image to align with the reference image, resulting in:
\begin{equation}
\label{eq:hypoplanes}
\mathcal{F}_{I_{Q}\rightarrow I_R}\big|_{\mathcal{Z}}=\mathcal{\phi}_{R}\Bigl[\mathcal{\phi}^{-1}_{Q}\Bigl(\mathcal{F}_{I_{Q}},\mathcal{Z}\Bigr)\Bigr].
\end{equation}
In practice, we utilize 2D rectifying grids that store the coordinates $\{x,y\}$ of feature correspondences, while the rectified features are obtained through bilinear interpolation at subpixel positions \cite{spatialtransformer}. Ultimately, the reference features along with a stack of query features are processed in the similarity decision step.  
\paragraph{Similarity computation.}
We evaluate the likelihood between per-pixel features using both cosine $\mathcal{S}^{cos}_{Z}$ or learned similarity $\mathcal{S}^{mlp}_{Z}$ metrics \cite{Chebbi_2023_CVPR,zbontar2016}. For a pair of feature maps $\mathcal{F}_{I_{R}}$ and $\mathcal{F}_{I_{Q}\rightarrow I_{R}}\big|_{Z}$, we express the resulting similarity at depth~$\mathcal{Z}$ as:
\begin{equation}
\label{eq:similarityPair}
\begin{array}{ll}
\mathcal{S}^{cos}_{Z}=\langle \mathcal{F}_{I_{R}},\mathcal{F}_{I_{Q}\rightarrow I_{R}}\big|_{Z} \rangle, \\
\mathcal{S}^{mlp}_{Z}=\mathcal{D}\Bigl(\mathcal{F}_{I_{R}},\mathcal{F}_{I_{Q}\rightarrow I_{R}}\big|_{Z}\Bigr),
\end{array}
\end{equation}
where $\langle.,.\rangle$ is the normalized dot product between embeddings at pixel location $\mathcal{P}$ (see Figure~\ref{fig:planesweep}); $\mathcal{D}\Bigl(.,.\Bigr)$ is the similarity score predicted by the decision MLP. In a general multi-view problem $(\textit{n} \geq 3)$, we select one reference image $\mathcal{I}_{R_{i}}$ and consider a batch of query images $\Bigl\{\mathcal{I}_{Q_{j}}, j \in \Bigl\{1,n\Bigr\}, i \neq j \Bigr\}$. The final similarity score is an aggregation of all similarities. In our experiments we rely on a simple average:
\begin{equation}
\label{eq:similarityaggregation}
\mathcal{S}^{agg}_{*_{Z}}=\frac{1}{n} \sum_{j} \mathcal{S}^{*}_{Z} \left(\mathcal{I}_{R_{i}},\mathcal{I}_{Q_{j}}\right)~, * \in \bigl\{cos,mlp\bigr\}
\end{equation}

\paragraph{Cost structure construction and regularization} The cost structure is defined by the geometry of the reference image, as well as by the depth search intervals. 
For every depth hypothesis $\mathcal{Z}$, its cost $\mathcal{C}^{agg}_{Z}$ is calculated as a function of the aggregated similarity score $\mathcal{S}^{agg}_{Z}$ in \Cref{eq:similarityaggregation}. The resulting cost volume/structure is regularized by semi-global matching~\cite{micmac, Hirsh}.
\begin{equation}
    \mathcal{C}^{agg}_{Z}=
    \begin{cases}
        1-\mathcal{S}^{agg}_{Z} & \text{if MLP is on}, \\
        \frac{1-\mathcal{S}^{agg}_{Z}}{2} & \text{if MLP is off}.
    \end{cases}
\end{equation}

\section{Datasets}
We conduct a comprehensive evaluation of \Ours~across several landscapes and acquisition geometries. To achieve this, we curate multi-view datasets with varied Ground Sampling Distance (GSD) and spanning diverse geographical locations. Encompassing predominantly urban areas, these datasets feature intricate scenes with narrow streets and repetitive building rooftops. 
\subsection{Ground truth depth map creation}
High quality training datasets are essential in training deep neural networks.
Self-supervised similarity learning for stereo matching relies on extracting pairs of matching and non-matching patches~\cite{zbontar2016} or locations~\cite{Chebbi_2023_CVPR} given an epipolar pair and a ground truth disparity map. A misalignment of images with respect to the ground truth LiDAR point cloud, whether spatial or temporal, can cause confusion within the network, potentially leading to issues such as model divergence or gradient collapse. {To ensure good co-registration accuracies in planimetry and altimetry, we pick GCPs on edges and planar surfaces. Then, the GCPs are leveraged to co-register the images to the Lidar data using bundle adjustment \cite{pierrot2012apero,rupnik2016refined}.}  To create ground truth depth maps, we first reproject all visible Lidar points into the geometry of the target image.
Next, inspired by \cite{biasutti2019}, we iteratively filter non-visible points by observing depths variability in a certain 2D neighbourhood of a pixel location in the image. This visibility analysis allows to discard almost all non-geometrically coherent points. To further refine the obtained sparse depth map, we perform closest-to-the-camera points selection to clear out the remaining points near buildings' outlines. 
Given a candidate point $P$, a vicinity neighbourhood $\mathcal{N}_{P}$ is defined based on the \textit{planar} $\mathcal{K}$-nearest neighbours. The visibility criterion for point $\mathcal{P}$ with depth $d_{p}$ can be computed based on \cite{biasutti2019} as:

\begin{equation}
\mathcal{V}=e^{\displaystyle{-\frac{(d_{p}-d_{p}^{min})^{2}}{(d_{p}^{max}-d_{p}^{min})^{2}}}}~;~0\le \mathcal{V}\le 1, 
\label{eq:biasutti}
\end{equation}
where $d_{p}^{min}$ and $d_{p}^{max}$ are minimum and maximum depths in the neighbourhood $\mathcal{N}_{p}$. Higher visibility scores describe points with lower depths, i.e closer to $d_{p}^{min}$. We set {empirically} $\mathcal{K}=27$ and the visibility threshold $\mathcal{V}_{th}=0.76$ to consider a point as visible by the camera. These parameters depend on point cloud density as well as depth variation, and require  careful tuning especially for coarser point clouds. 
Once the occluded points removed, we densify the ground truth depth maps using Delaunay triangulation as presented in \cite{Chebbi_2023_CVPR}. Densification provides additional training samples and enhances the accuracy metrics' reliability by incorporating structural details of objects. 
\subsection{Training}
We train the proposed \Ours~models (as explained in \Cref{subsec:prelimin} nd ~\cite{Chebbi_2023_CVPR}) using a benchmark dataset for aerial imagery introduced by \cite{wuteng}. This dataset comprises epipolar image pairs from diverse locations, including Enschede, Vaihingen and Dublin~\cite{dublincity}. We use a total of 34,612 image pairs to fully train the models. The size of each image is 1024pix $\times$ 1024pix and every batch contains 16 image pairs. To enhance generalization, the network is never exposed to the entire 1024x1024 region. Instead, during each iteration, a distinct subregion of the reference image is presented to the network. The origin of the subregion is randomly selected, while the sizes vary among three options: 512x512, 640x640, or 768x768 pixels. The inclusion of the latter size variation helps make the network more resilient to different patch sizes. To maintain consistency, the corresponding query image is cropped to match the height of the reference sub-region, while preserving the entire width of the image. This ensures to capture the per-pixel embeddings along the epipolar line.
Note that our training is self-supervised as ground truth disparities are only utilized to identify \textit{matching} and \textit{non-matching} pixel pairs but do not contribute to the final training loss function. 
\subsection{Testing}
\begin{figure}[t!]
    \centering
    \includegraphics[height=0.247\textwidth,width=\textwidth]{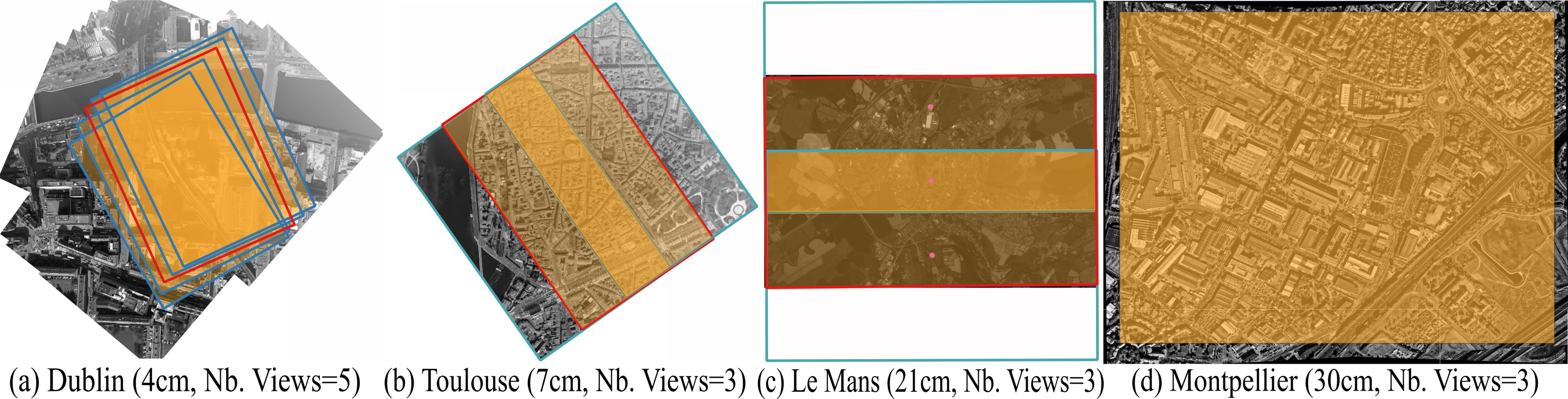}
    \caption{\textbf{Datasets.} Four test datasets representing \textit{in-distribution} (Dublin, Toulouse) and \textit{out-of-distribution} (Le Mans, Montpellier) scenarios. The highlighted zones indicate overlapping areas, with color intensity increasing as the number of overlapping images grows. In Montpellier, two tri-stereo acquisitions cover the entire region of interest.}
    \label{fig:datasetsAndViews}
\end{figure}

During training our networks see only high resolution aerial images. However, during testing, we assess their performance on both aerial and satellite acquisitions. By extending the test set to satellite imagery, we explore the potential and limitations of both our training strategy, as well as the resilience of the features under informational content drop due to the lower signal-to-noise ratio inherent to satellite imagery. We refer to our datasets throughout the paper as \textit{Dublin}, \textit{Toulouse}, \textit{Le Mans} and \textit{Montpellier}. Dublin (4cm GSD) and Toulouse (8cm GSD) are considered \textit{within-distribution} datasets, whereas Le Mans (21cm GSD) and Montpellier (30cm GSD) are \textit{out-of-distribution} datasets (see \Cref{fig:datasetsAndViews}). Notice that even though Dublin appears in both the training and testing datasets, the two sets are disjoint, with no images appearing in both. 

\subsection{Data normalisation}
Radiometric normalization within deep neural networks speeds up the calculations while stabilizing the model convergence. Our multi-city test set is comprised of datasets with different radiometric distributions. Lack of prior radiometric preprocessing can significantly impact model performance. We propose normalizing image patches using region-specific statistics. Precisely, pre-processing involves scaling all radiometries to $[$0,1$]$ range, followed by normalization using the mean and standard deviation of radiometries. During training, these statistics are computed over the entire training set. During testing, the radiometric distribution of the dataset may have a unique center and spread, differing from the training set. Therefore, mean and standard deviations are recalculated for the dataset under test. Both training and testing utilize 8-bit grayscale images.

\section{Numerical Experiments}

\subsection{Evaluation}
The computed depth maps are evaluated against ground truth Lidar as well as against competitive dense image matching algorithms including MC-CNN \cite{zbontar2016} (using our training strategy), PSMNet \cite{PSMNet}, RAFT-Stereo~\cite{raftstereo} and MicMac \cite{micmac}.
{Our PSMNet model was trained on the Aerial Stereo Dense Matching Benchmark \cite{wuteng}.} {For RAFT-Stereo we used the improved vanilla model denoted as iRaftStereo\_{RVC} that ranked $2^{nd}$ in the Robust Vision Challenge~(RVC) 2022. The model has been trained on a mixture of large synthetic and realistic stereo datasets including SceneFlow\cite{sceneflow},~TartanAir\cite{tartanair2020iros} and CreStereo\cite{crestereo}. More than 577,545 stereo pairs are used for training.  Given the training dataset size, we excluded the option of retraining RAFT-Stereo on our stereo datasets which altogether are over 15 times smaller.} 

\begin{figure}[H]
    \centering
\includegraphics[width=1.\textwidth]{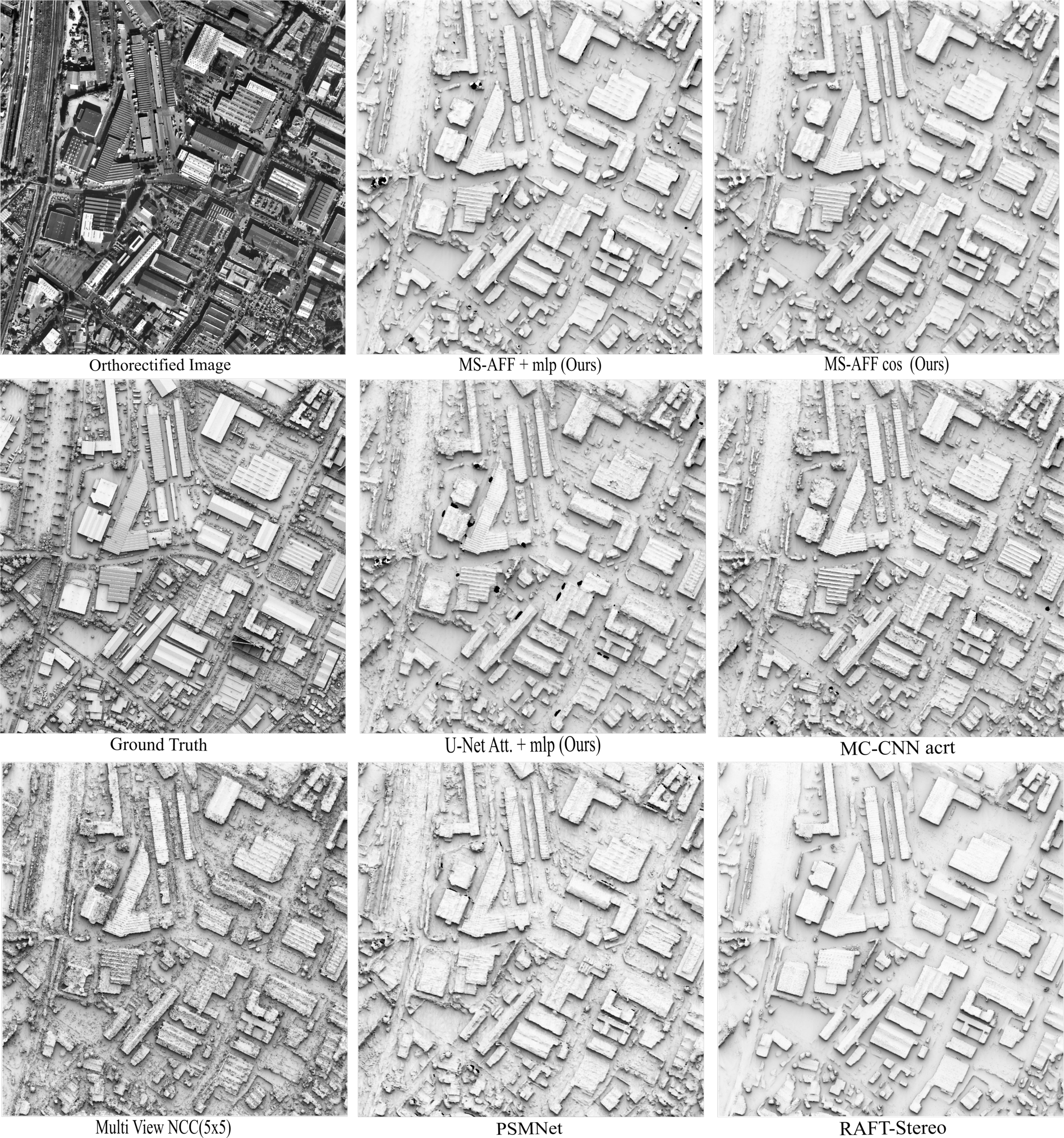}
    \caption{\textbf{Evaluation Metrics Limitations.} Which model is the best?  Although statistically speaking, the standard NCC and RAFT-Stereo  yield the best results (see \Cref{tab:statsmontpellier}), rendered buildings are either noisy with fuzzy outlines or too smoothed. Conversely, MS-AFF renders  noise-free and "meaningful" objects with neat borders. Clearly, the error histograms based accuracy analysis reaches a saddle point with higher GSDs -- small noisy details that are recovered by NCC but neglected by \Ours~ contribute to the overall statistics without describing semantically meaningful objects. }
    \label{fig:groundtruth_issue}
\end{figure}

We adopt several quantitative metrics to reveal their accuracy, robustness and completeness. Precisely, we use the mean absolute error (MAE, denoted as $\mu$), absolute error histograms, the standard deviation errors ($\sigma$), normalized median absolute deviation (NMAD), and cumulative errors across n-times the ground sampling distance (GSD), represented as $D_{xGSD}$ and expressed in $\%$. The error is defined as the deviation from ground truth depth over all inlying pixels (error $<$1.5 m). Completeness is the ratio of inlier pixels to total pixels. We also provide qualitative results in form of grayshaded depth maps (see \Cref{fig:dublinresultsmlp,fig:dublinresults,fig:epipolarhomographictoulouseurban1,fig:gsd20cm,fig:gsd20cm_mccnandcorrel,fig:gsd30cm}). We also point out the limits of the ground truth Lidar due to non synchronous image-Lidar acquisition in Figure~\ref{fig:gsd30cm}, and unbalanced spatial sampling between both modalities that induces a semantic mismatch in \Cref{fig:groundtruth_issue}. Runtimes are given in \Cref{tab:runtimeEfficiency}.

\subsection{Implementation details}
\Ours~is integrated within MicMac's~\cite{micmac,epipolarresampling} dense matching pipeline. The pipeline follows a hierarchical framework with finer image resolutions being refined based on depth predictors computed at lower image resolutions.   
Our models can be activated within this hierarchical scheme. Deep features are more distinctive when spatial resolutions increase (see \Cref{fig:explain_why_ncc_good_low_resolution}), and we observed that activating the tested deep models from $\frac{\mathcal{R}}{4}$ yields sufficiently robust features to construct accurate depth predictors. 
{Consequently, the cost structure is bounded to previous depth predictors with lower memory requirements~(see Table~\ref{tab:runtimeEfficiency}). However, unlike the tested end-to-end networks, our models process the reference and query images separately due to their differing sizes, which may slightly increase the processing time. All in all, our approaches process a 512 $\times$ 512 epipolar pair in less than 1s. Without holding the whole feature cost volume in memory, our method enables processing multiple pairs in parallel. }

We distinguish between multi-view and multi-view stereo, also referred to as \textit{fusion}. Multi-view stereo involves generating several two-view depth maps in image geometry. These maps are then aligned in a common reference frame in ground geometry, followed by a fusion process, as detailed in~\cite{rupnik20183d}. For all multi-view scenarios, given a set of camera views, we select one view as reference. When using epipolar priors, the epipolar features are warped back to the initial image geometry with grid sampling~\cite{spatialtransformer}. When using homography prior, reference image features are computed in original geometry, whereas the query features are computed over images resampled with a homographic warping. De-warping is therefore conducted exclusively on query features which is faster than the equivalent epipolar processing.  To explore depth hypotheses, geometry-aware features are bilinearly resampled based on \textit{2-D} warping grids (see Figure~\ref{fig:planesweep}). Finally, at a given depth hypothesis, the interpolated features are compared with the cosine or learned MLP functions, and passed to the cost structure. The cost is ultimately regularized with SGM.  Regularization of the costs provided by \Ours~is 2 orders of magnitude smaller than that of MC-CNN and NCC.

\begin{table}[t]
    \centering
    \begin{tabular}{p{2.5cm}cccc}
    \Xhline{2\arrayrulewidth}
       \multirow{2}{*}{\textbf{Method}}  &  \multicolumn{3}{c}{\textbf{Runtime~(s)} GPU @ 1305 MHz} & \multirow{2}{*}{\textbf{GPU Usage~(MB)}}\\
       \cline{2-4} & Feature extraction & Similarity computation + SGM & Total & \\
       \hline
       MS-AFF~(Ours)  &  0.65 & 0.24  & 0.89 & 824\\
       U-Net~(Ours)   &  0.42 & 0.24  & 0.66 & 854\\
       MC-CNN~\cite{zbontar2016} & 0.2 & 0.26 & 0.46 & 592\\
                   PSMNet~\cite{PSMNet}  & - & - & 0.65 & 2932\\
       RAFTStereo~\cite{raftstereo} & - & - & 0.78 & 1756\\
       \hline
    \end{tabular}
    \caption{\textbf{Models Runtime.} Runtime is computed on epipolar rectified images of size 512 $\times$ 512. For MS-AFF, 4 multi-resolution images are passed to the network. This justifies the small increase in feature extraction runtime compared to \Unet. Note that our models take left and right images separately, as sizes may vary given disparity search intervals. PSMNet and RAFT-Stereo jointly infer on concatenated pairs. }
    \vspace{-0.5cm}
    \label{tab:runtimeEfficiency}
\end{table}

\subsection{Discussion}
\vspace{0.2cm}
\subsubsection{The influence of multi-views}
\begin{figure}[H]
    \centering
\includegraphics[height=0.25\textwidth,width=\textwidth]{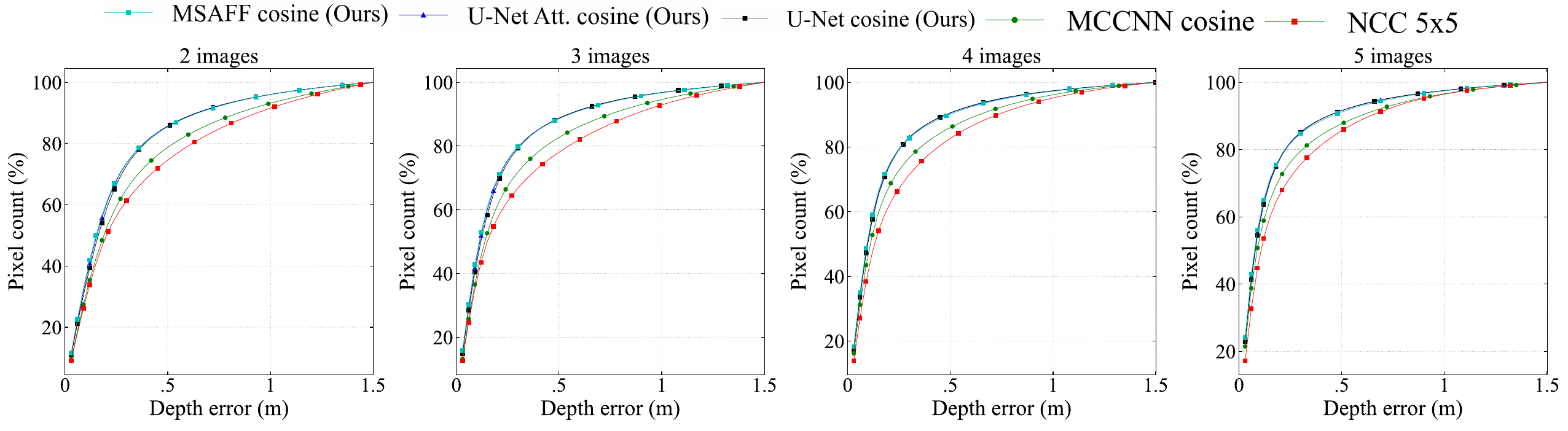}
\includegraphics[height=0.25\textwidth,width=\textwidth]{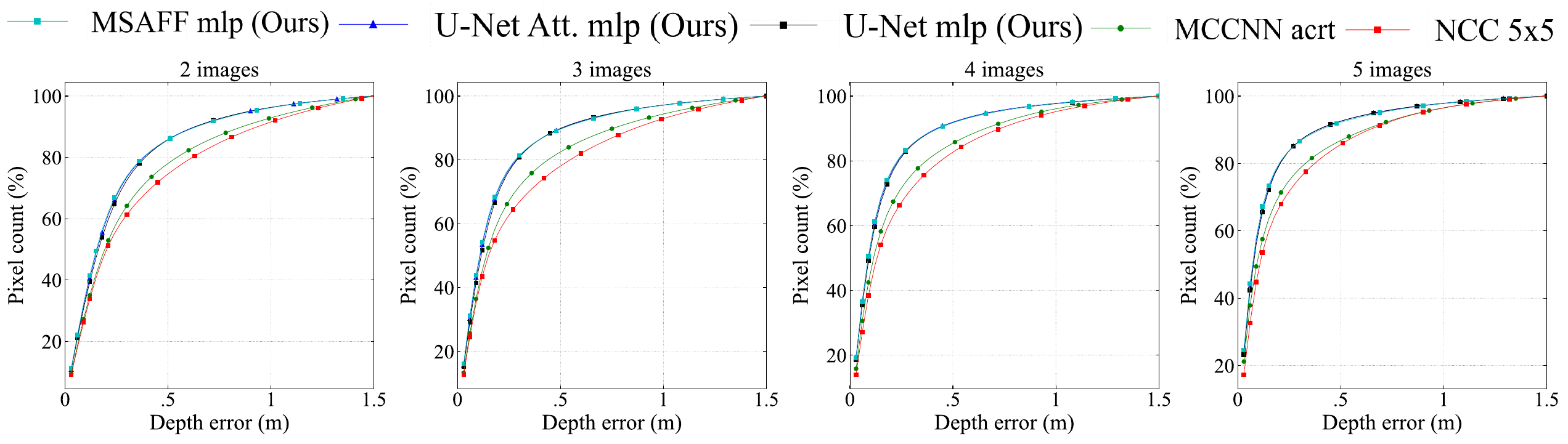}
    \caption{\textbf{Multi-view -- Absolute Error Histogram.} In-distribution Dublin dataset: Absolute error cumulative histograms computed for several \Ours~variants (MS-AFF, U-Net, U-Net Att.), MC-CNN trained with our framework and normalized cross-correlation NCC. The steepness of all curves grows with the increasing number of views. MC-CNN and NCC perform worst, while the neural networks perform best and comparably among them. The MLP proves beneficial for scenarios with 3-views and more.
    } 
    \label{fig:dublinhistogramsepip}
\end{figure}

Aggregating pairwise learnt similarity reduces the portion of depth outliers (see \Cref{fig:dublinresultsmlp,fig:dublinresults}) while preserving the buildings' rooftops details and the buildings' outlines. As illustrated in Table~\ref{tab:statsdublin}, the accuracy increases across all models as more views are added. The best performing models on both \textit{in-distribution} Dublin (Table~\ref{tab:statsdublin}, Figure~\ref{fig:dublinhistogramsepip}) and \textit{out-of-distribution} Le Mans (Table~\ref{tab:statslemans}) datasets are MS-AFF and U-Net Attention. {Unlike PSMNet that exhibits a performance drop with increasing GSDs~\cite{Chebbi_2023_CVPR}, RAFT-Stereo yields decent reconstructions without further training~(see Table \ref{tab:statslemans}). However, both models' predictions neglect high frequency details and render smooth DEMs~(see \Cref{fig:gsd20cm_mccnandcorrel,fig:gsd20cm}).} Compared to image matching relying on local neighbourhood, i.e. the traditional NCC or MC-CNN, \Ours~is consistently better ranked and the most complete (see Table~\ref{tab:completeness}). In Dublin dataset, MS-AFF \textit{cosine} surpasses MC-CNN \textit{cosine} by 7.6 $\%$ under 2-views conditions for $D_{3\times\textit{GSD}}$ metric. This gap jumps up to 8.1 $\%$ for 3-views configuration. For $\{4,5\}$-views, it  maintains a minimum of 6$\%$ increase. This shows that increasing views provides overall more depth consistency for all correlators. However, in spite of MC-CNN being trained on the same datasets as \Ours, even in multi-view configurations its limited receptive field introduces a lack of distinctiveness, especially over textureless regions.

\begin{figure}[H]
\vspace{-0.6cm}
    \centering
    \includegraphics[width=0.8\textwidth]{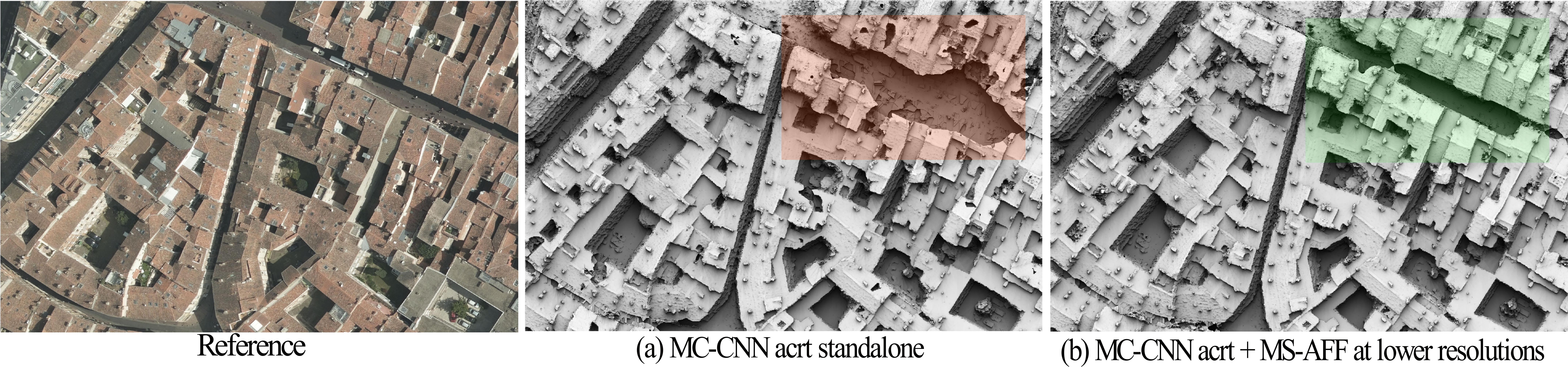}
    \caption{\textbf{Ambiguities Induced by Large Depth Search Intervals Experiment.} MC-CNN acrt~\cite{zbontar2016} fails to reconstruct $\frac{\mathcal{R}}{8}$ and $\frac{\mathcal{R}}{4}$ depth levels. Errors propagate to high resolution depths with MC-CNN standalone. To compensate for these depth prediction errors \textbf{(a)}, we rely on MS-AFF features to robustly provide a decent depth prior and recover wrong depths \textbf{(b)}.}
    \label{fig:mccnnsfailsatcomplexurbanwithhomography}
\end{figure}

\begin{figure}[H]
\vspace{-0.6cm}
    \centering
    \includegraphics[width=0.8\textwidth]{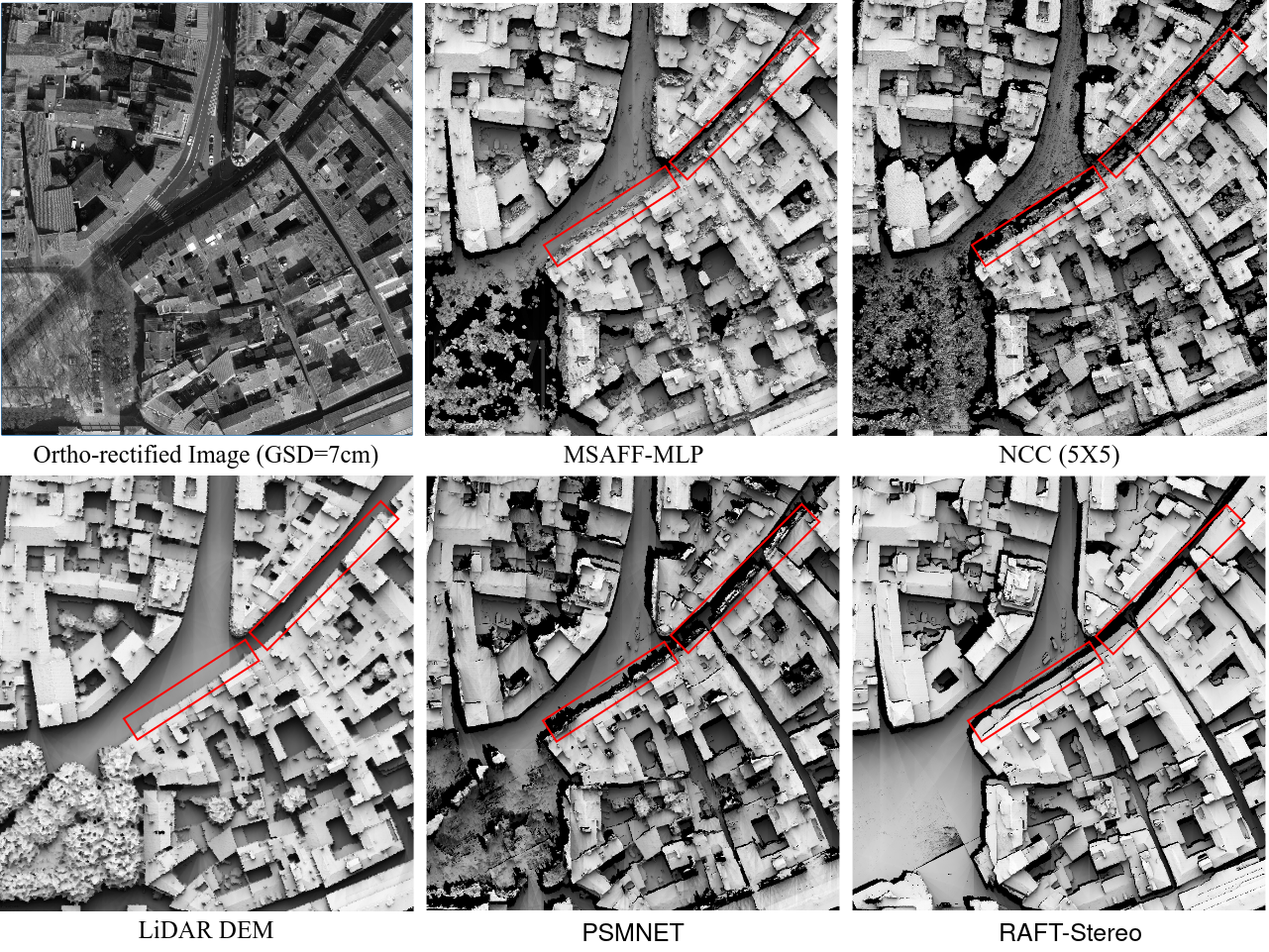}
    \caption{\textbf{Large $\frac{B}{H}$(=0.43) Experiment. } DEMs generated by different models. \Ours~(MSAFF-MLP) yields unstructured reconstructions in occluded areas while maintaining high frequency details together with the NCC-based approach. MSAFF-MLP fails to recover narrow roads as no matches are likely to occur in these regions. RAFT-Stereo~\cite{raftstereo} and PSMNet~\cite{PSMNet} render smoothed buildings rooftops and miss high frequency details while artificially interpolating reconstructions in occluded areas and narrow roads.}
    \label{fig:largebhexperiment}
\end{figure}

Moving from a stereo to a tri-stereo configurations significantly boosts the accuracy across the majority of models. {The $\frac{B}{H}$ ratios for $\{2,3,4,5\}$-views are $\{0.06,0.06,0.12,0.13\}$ respectively~(see \Cref{tab:statsdublin}, \Cref{fig:dublinhistogramsepip,fig:dublinresultsmlp,fig:dublinresults})}. 
Query images with low $\frac{B}{H}$ produce highly concordant similarities and little or no occlusion artifacts. Increasing the $\frac{B}{H}$ ratio introduces more pronounced perspective deformations and more severe occlusions to which a simple similarity averaging is not adapted. This explains the deterioration of facades' reconstructions beginning with the 4-view setting, in particular for the cosine similarity (see Figure~\ref{fig:dublinresults}). Adding a more sophisticated similarity score aggregation scheme or simply visibility masks can resolve the issue, but is out of the scope of this work. The experiment with larger $\frac{B}{H}$(=0.43) (see \Cref{fig:largebhexperiment}) reveals a performance drop for all tested models, including~\Ours. 
The latter performs poorly because it was not trained on such large $\frac{B}{H}$ configurations~\cite{Chebbi_2023_CVPR} and because it was supervised to increase dissimilarity in occluded areas. Conversely, the end-to-end regression networks (PSMNet, RAFTStereo) are supervised even in occluded areas leading to smooth interpolations in those regions.  Furthermore, although our multi-resolution depth prediction accelerates reconstruction, but also propagates matching errors from lower resolutions and oftentimes prevents~\Ours~from capturing the bottom of narrow streets~(see \Cref{fig:mccnnsfailsatcomplexurbanwithhomography}).

\begin{table}[H]
    \centering
    \begin{adjustbox}{max width=0.98\textwidth}
    \begin{tabular}{@{}l@{}>{\raggedleft\arraybackslash}lcccccc|ccccccc@{}}
    \Xhline{2\arrayrulewidth}
    & \textbf{Configuration} & \multicolumn{6}{c}{\textbf{2 \text{images}}} & \multicolumn{6}{c}{\textbf{3 \text{images}}} \\
    & Architecture & $\mu$$\downarrow$ & $\sigma$$\downarrow$ & $\textit{NMAD}$$\downarrow$ & $D_{1.5\times\textit{GSD}}$$\uparrow$ & $D_{3\times\textit{GSD}}$$\uparrow$ & $D_{5\times\textit{GSD}}$$\uparrow$ & $\mu$$\downarrow$ & $\sigma$$\downarrow$ & $\textit{NMAD}$$\downarrow$ & $D_{1.5\times\textit{GSD}}$$\uparrow$ & $D_{3\times\textit{GSD}}$$\uparrow$ & $D_{5\times\textit{GSD}}$$\uparrow$ \\
    \Xhline{2\arrayrulewidth}
         & MS-AFF \text{cosine} & 0.25 & 0.29 & \textbf{0.1} & \textbf{22.6} & \textbf{49.9} & \textbf{62.3} & 0.21 & 0.27 & \textbf{0.08} & \textbf{30.3} & \textbf{60.7} & \textbf{71.2}\\ 
         & U-Net \text{cosine} & 0.26 & 0.28 &  0.11 & 21.1 &47.2& 60.1 & 0.21 & 0.27 & 0.08 & 28.5 &58.3 &69.9\\
         & U-Net Att. \text{cosine}  & \textbf{0.25} &  \textbf{0.28} & 0.1 &21.7 & 48.9 & 61.7 & \textbf{0.21} & \textbf{0.27} & 0.08 & 29.3 &59.9 & 70.9\\
         & MCCNN \text{cosine}~\cite{zbontar2016}  &  0.32 & 0.34 & 0.13 & 18.7 & 42.3 & 53.7 & 0.27 & 0.32 & 0.1 & 25.7 & 52.6 & 62.8\\
         \hline
         & MS-AFF \text{mlp} & \textbf{0.25} & 0.28 & \textbf{0.1} & 22.1 &\textbf{49.5} & \textbf{62.1} & \textbf{0.2} & 0.26 & \textbf{0.07} &\textbf{31.3} &\textbf{62.2} & \textbf{72.9}\\
         & U-Net \text{mlp}  & 0.25 & 0.28 & 0.1 & 21.1 & 47.2 & 59.8 & 0.2 & \textbf{0.25} & 0.08 & 29.2 &59.9 & 71.5\\
         & U-Net Att.  \text{mlp}  & 0.25 & \textbf{0.28} & 0.1 & \textbf{21.6} & 48.6 & 61.4 & 0.2 & 0.26 & 0.07 & 30.3  & 61.5 & 72.3\\ 
         & MCCNN \text{acrt}~\cite{zbontar2016}   & 0.32 & 0.34 & 0.14 & 18.6 & 41.8 & 52.9 & 0.27 & 0.32 & 0.1 & 25.7& 52.4 & 62.5\\ 
         \hline
         & NCC $5\times5$~\cite{micmac}  & 0.35 & 0.36 & 0.15 & 17.9 & 40.6 & 51.2 & 0.3 &  0.35 &  0.12 & 24.6 & 49.8 & 58.6\\
         \Xhline{2\arrayrulewidth}
         \\
         \Xhline{2\arrayrulewidth}
         & \textbf{Configuration} & \multicolumn{6}{c}{\textbf{4 \text{images}}} & \multicolumn{6}{c}{\textbf{5 \text{images}}} \\
    & Architecture & $\mu$$\downarrow$ & $\sigma$$\downarrow$ & $\textit{NMAD}$$\downarrow$ & $D_{1.5\times\textit{GSD}}$$\uparrow$ & $D_{3\times\textit{GSD}}$$\uparrow$ & $D_{5\times\textit{GSD}}$$\uparrow$ & $\mu$$\downarrow$ & $\sigma$$\downarrow$ & $\textit{NMAD}$$\downarrow$ & $D_{1.5\times\textit{GSD}}$$\uparrow$  & $D_{3\times\textit{GSD}}$$\uparrow$ & $D_{5\times\textit{GSD}}$$\uparrow$\\
    \Xhline{2\arrayrulewidth}
         & MS-AFF \text{cosine} & 0.19 & 0.25 & 0.06 & \textbf{35.0} & \textbf{66.4} & \textbf{75.5} & 0.16 & 0.24 & 0.05 & \textbf{43.0} & \textbf{71.1} & 78.7\\ 
         & U-Net \text{cosine}  & 0.19 & 0.25 & 0.06 & 33.5 & 65.2 &75.0 & 0.17 & 0.24 & 0.05 & 41.3 & 70.2 & 78.4\\
         & U-Net Att. \text{cosine}  & \textbf{0.18} & \textbf{0.25} & \textbf{0.06} & 34.5 & 65.8 & 75.4 & \textbf{0.16} & \textbf{0.23} & \textbf{0.05} & 42.2 & 70.7 & \textbf{78.8}\\
         & MCCNN \text{cosine}~\cite{zbontar2016} & 0.23 & 0.29 & 0.08 & 31.2 & 59.7 & 68.8 & 0.21 & 0.28 & 0.05  & 38.7 & 64.8 & 72.7\\   
         \hline
         & MS-AFF \text{mlp} & \textbf{0.17}  & \textbf{0.24}  & \textbf{0.06} & 36.5 & \textbf{68.8} & \textbf{78.0} & 0.16 & 0.24 & \textbf{0.05} & \textbf{44.3} & \textbf{73.4} & \textbf{80.8}\\
         & U-Net \text{mlp}  & \textbf{0.17} & \textbf{0.24} & \textbf{0.06}  & 35.5 & 67.2 & 77.1 &  0.16 & \textbf{0.23} & \textbf{0.05} & 42.3 & 72.1 & 80.4\\
         & U-Net Att.  \text{mlp}  & \textbf{0.17} & \textbf{0.24} & \textbf{0.06} & \textbf{36.7} & 68.2 & 77.6 & \textbf{0.15} & \textbf{0.23} & \textbf{0.05} & 43.5 & 72.7 & 80.6 \\ 
         & MCCNN \text{acrt}~\cite{zbontar2016} & 0.23 & 0.29  & 0.08 & 30.6 & 58.2 & 67.4 & 0.21 & 0.28 & 0.07 & 37.8 & 63.4 & 71.4\\ 
         \hline
         & NCC $5\times5$~\cite{micmac}   & 0.26 & 0.31 & 0.1 & 27.1 & 54.1 & 63.0 & 0.22 & 0.29 & 0.08 & 32.6 & 59.8 & 67.9\\
         \Xhline{2\arrayrulewidth}
    \end{tabular}
    \end{adjustbox}
    \caption{\textbf{Multi-view -- Accuracy.} Dublin dataset: Error metrics computed for several \Ours~variants (MS-AFF, U-Net, U-Net Att.), MC-CNN trained with our framework and normalized cross-correlation NCC. 
    Augmenting the number of views improves all metrics. The MLP similarity outperforms the cosine similarity everywhere except for the 2-view configuration. MS-AFF places among the best preforming variants of deep similarity networks. See Figures \ref{fig:dublinresultsmlp}~\&~\ref{fig:dublinresults} for visual quality assessment. }
    \label{tab:statsdublin}
\end{table}

\begin{table}[H]
\centering
\begin{adjustbox}{max width=0.7\textwidth}
\begin{tabular}{@{}l@{\hskip 0.5cm}>{\raggedleft\arraybackslash}cccccccc@{}}
    \Xhline{2\arrayrulewidth}
    \textbf{Configuration} & \multicolumn{2}{c}{\textbf{2 \text{images}}} & \multicolumn{2}{c}{\textbf{3 \text{images}}} & \multicolumn{2}{c}{\textbf{4 \text{images}}} & \multicolumn{2}{c}{\textbf{5 \text{images}}}\\
    Architecture & Epip. & Hom. & Epip. & Hom. & Epip. & Hom. & Epip. & Hom.\\
    \hline
    MS-AFF cosine & 86.5 & 86.5 & 87.9 & 87.9 & 88.0 & - & 88.1 & - \\
    U-Net cosine  &  87.0 & 87.0 & 89.0  & 88.9 & 89.1 & - & 89.2 & \\
    U-Net Att. cosine & 87.3 & 87.2 & 89.1 & 88.8 & 89.1 & - & 89.1 & - \\
    MC-CNN cosine~\cite{zbontar2016} & 81.9 & 81.6 & 83.8 & 83.4 & 85.8 & - & 86.2 &  - \\
    \hline
    MS-AFF mlp & 86.6 & 86.2 & 88.8 & 87.7 & 88.7 & - & 88.7 & - \\
    U-Net mlp  & 87.4 & 87.3 & 89.7 & 89.7 & 89.8 & - & 90.1 & - \\
    U-Net Att. mlp & 87.6 & 87.5 & 90.0 & 89.8 & 89.8 & - & 90.2 & - \\
    MC-CNN acrt~\cite{zbontar2016} & 82.0 & 81.3 & 85.1 & 84.3 & 88.5 & - & 88.9 & - \\
    \hline
    NCC $5\times5$~\cite{micmac} & 82.5 & - & 86.6 & - & 89.3 & - & 90.3 & - \\
    \Xhline{2\arrayrulewidth}
\end{tabular}
\end{adjustbox}
  \caption{\textbf{Multi-view -- Completeness.} In-distribution Dublin dataset: The completeness (in $\%$) increases with the growing number of views and is higher for scenarios with MLP than scenarios with cosine similarities. \Ours~with U-Net variant retains most of the pixels. The epipolar prior performs better than the homography prior, especially if more views are used.
  }
    \label{tab:completeness}
\end{table}

\begin{table}[H]
    \centering
    \begin{tabular}{l@{}lcccccc}
    \Xhline{2\arrayrulewidth}
    & Architecture & $\mu$$\downarrow$ & $\sigma$$\downarrow$ & $\textit{NMAD}$$\downarrow$ & $D_{1\times\textit{GSD}}$$\uparrow$ & $D_{2\times\textit{GSD}}$$\uparrow$ & $D_{3\times\textit{GSD}}$$\uparrow$\\
    \Xhline{2\arrayrulewidth}
         & \colorbox{blueblue}{MS-AFF \text{mlp} \textit{epip}} & 0.33 & 0.31 & 0.15 & 44.4 & 71.9 & 85.0\\
         & \colorbox{blueblue}{MS-AFF \text{mlp} \textit{hom}} & 0.33 & 0.31 & 0.15   & 45.4& 72.6  & 85.1\\
         & \colorbox{redred}{MS-AFF \text{mlp} \textit{fusion}} & 0.35 & 0.31 & 0.16  & 41.7 & 69.6 & 83.9\\
         \hline
         & \colorbox{blueblue}{U-Net \text{mlp}  \textit{epip}} & 0.34 & 0.31 & 0.16 & 43.4 & 70.7 & 84.2\\
         & \colorbox{blueblue}{U-Net \text{mlp}  \textit{hom}} & 0.34 & 0.31 & 0.15 & 45.2 & 71.9 & 84.6 \\
         & \colorbox{redred}{U-Net \text{mlp}  \textit{fusion}} & 0.35 & 0.31 & 0.16 & 41.3 & 69.3 & 83.8\\
         \hline
         & \colorbox{blueblue}{U-Net Att.  \text{mlp}  \textit{epip}} & 0.34 & 0.31 & 0.16 & 44.0 & 71.6 & 84.8\\ 
         & \colorbox{blueblue}{U-Net Att.  \text{mlp}  \textit{hom}} & 0.33 & 0.31 & 0.15  & 45.2  & 72.5  &  84.8\\
         & \colorbox{redred}{U-Net Att.  \text{mlp}  \textit{fusion}} & 0.35 & 0.31 & 0.16 & 42.2 & 70.0  & 84.3\\
         \hline
         & \colorbox{blueblue}{MC-CNN \text{acrt}~\cite{zbontar2016}  \textit{epip}} & 0.39 & 0.33 & 0.19 & 37.6 & 64.1 & 79.8\\ 
         & \colorbox{blueblue}{MC-CNN \text{acrt}~\cite{zbontar2016}  \textit{hom}} & 0.36 &  0.32 &  0.17 & 42.5 & 69.0 & 82.7\\ 
         & \colorbox{redred}{MC-CNN \text{acrt}~\cite{zbontar2016}  \textit{fusion}} & 0.45 & 0.34 & 0.22 & 29.8 & 55.8 & 74.2 \\ 
         \hline
         & \colorbox{blueblue}{NCC $5\times5$~\cite{micmac} \textit{multiview}} & 0.39 & 0.34 & 0.18 & 39.2 & 65.3 & 79.8\\
         & \colorbox{redred}{PSMNet~\cite{PSMNet}} & 0.34 & 0.28 & 0.16 & 38.6 & 69.3 & 85.6 \\
         & \colorbox{redred}{RAFT-Stereo~\cite{raftstereo}} & 0.34 & 0.32 & 0.16 & 44.8 & 71.9 & 84.4 \\
         \Xhline{2\arrayrulewidth}
    \end{tabular}
    \caption{\textbf{\colorbox{blueblue}{Multi-View} or \colorbox{redred}{Multi-view Stereo} -- Three-View Experiment.} Out-of-distribution Le Mans dataset: Multi-view (MV) \OursBase~variants with different geometry priors (\textit{epip, hom}) are compared with the multi-view stereo (MVS) equivalents. MVS run the stereo matching in image geometry with every possible pair of images, followed by a fusion of the three depth maps in ground geometry \cite{rupnik20183d}. MV consistently outperforms MVS for both geometric priors. Unlike within the \textit{in-distribution} datasets, in this \textit{out-of-distribution} dataset the homography prior is the best performing variant.}
    \label{tab:statslemans}
\end{table}

\subsubsection{Multi-view or multi-view stereo} 
Although~\OursBase~yield reliable similarity measures given a pair of images in epipolar geometry~\cite{Chebbi_2023_CVPR}, relying on additional views and thereby incrementing the per-candidate depth likelihood values improves the resulting similarity metrics. We compare this approach, referred to multi-view (or \textit{early fusion}), with multi-view stereo (or \textit{late fusion}). \textit{Early fusion} votes for a given depth by capitalising on learnt similarities' agreement among multiple views and modifies the cost volume accordingly. \textit{Late fusion}, on the other hand, accumulates depth information coming from different views \textit{a posteriori}, at the surface level, by combining per-pixel candidate depth candidates and their prediction confidence~\cite{rupnik20183d}. We show that both epipolar and homography priors for \textit{early} learnt similarity \textit{fusion} consistently outperform \textit{late} surface \textit{fusion}, i.e multi-view stereo (see Table~\ref{tab:statslemans}).

\subsubsection{Cosine or MLP similarities}

\begin{figure}[H]
    \centering
    \includegraphics[height=1.18\textwidth,width=0.9\textwidth]{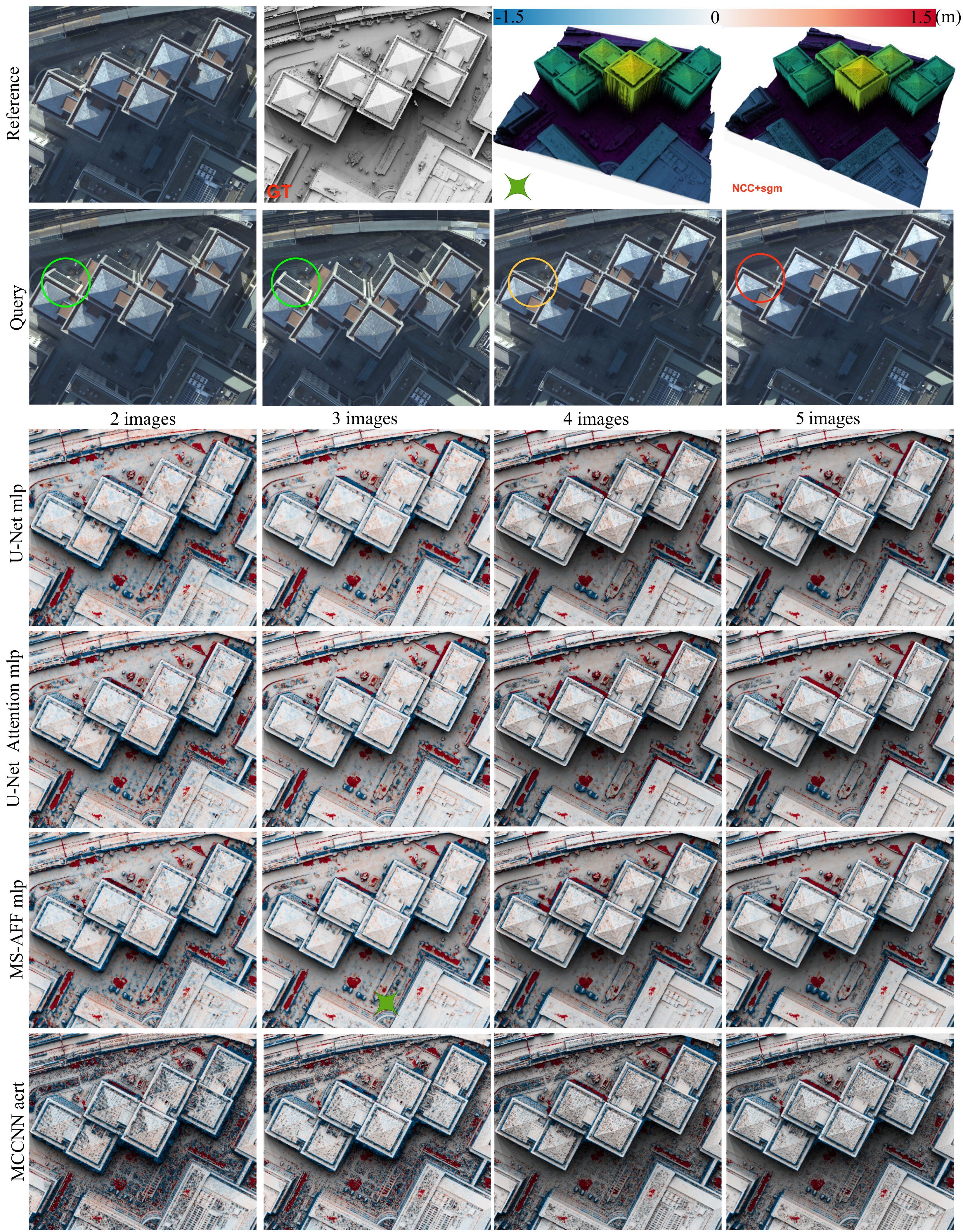}
    \vspace{-0.2cm}
    \caption{\textbf{Depth Map Visual Quality as Function of Number of Views -- MLP similarity}. In-distribution Dublin dataset: \Ours~depth maps reconstructed with incremental number of views along columns. Column 1 is the result of an image pair~(Reference,Query). Here, the MLP is activated and outputs learnt similarities. \Ours~yield noiseless reconstructions as the number of view increases on non-occluded areas. As facades become occluded beginning with 3-views, our models only partially recover the facade surface. MC-CNN \textit{acrt} reconstructions are noisy even with added views. We report a 3D view of the best and worst models, i.e MS-AFF mlp and multi-view cross correlation. {The $\frac{B}{H}$ ratios are $\{0.06,0.06,0.12,0.13\}$  for $\{2,3,4,5\}$-views respectively.}}
    \label{fig:dublinresultsmlp}
\end{figure}

\begin{figure}[H]
    \centering
    \includegraphics[height=1.06\textwidth,width=\textwidth]{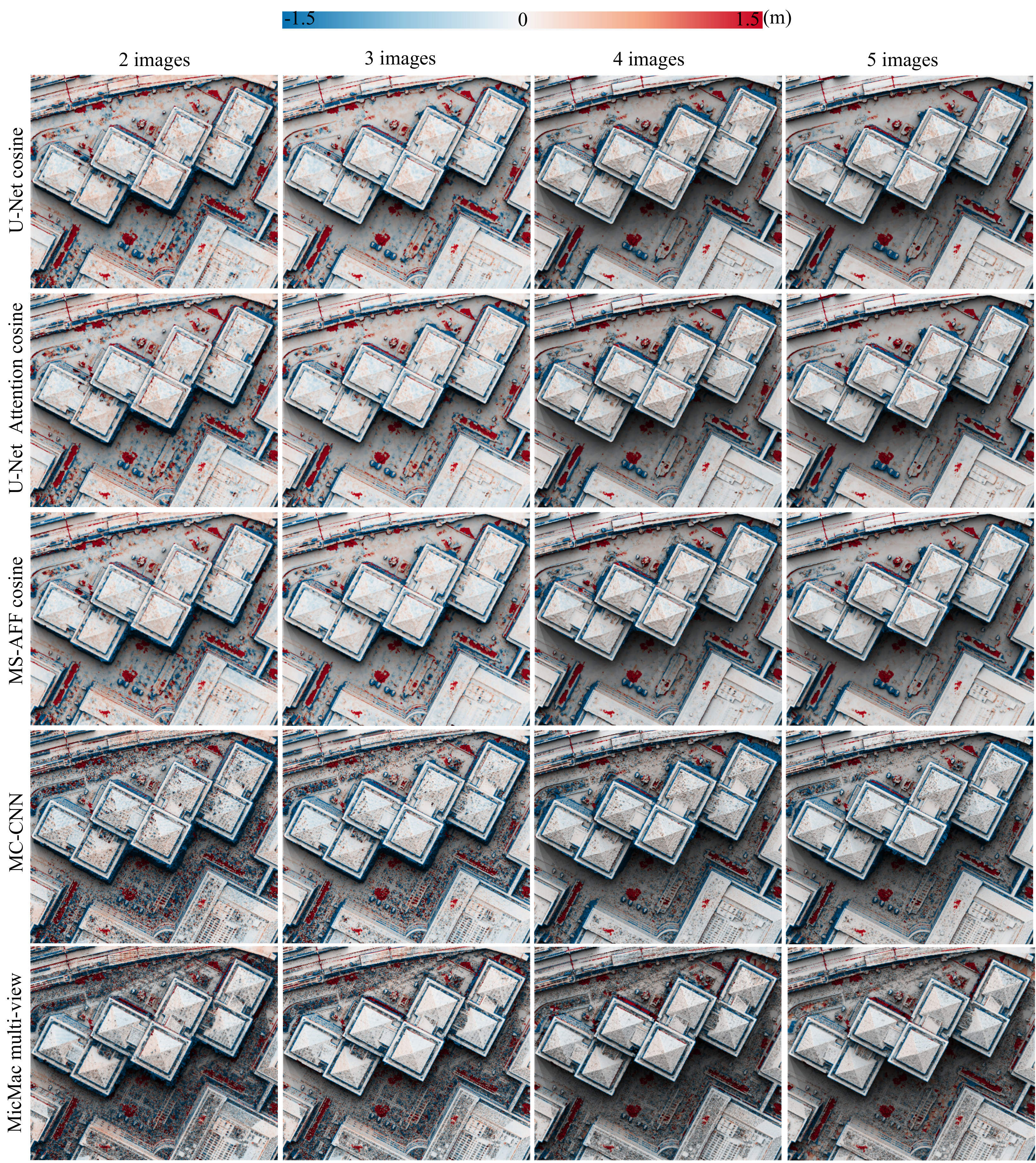}
    \caption{\textbf{Depth Map Visual Quality as Function of Number of Views -- Cosine similarity.} In-distribution Dublin dataset: Cosine-based surface reconstructions: Views are displayed in \Cref{fig:dublinresultsmlp}. \Ours~features are sufficiently expressive at this GSD level (4cm) and reduce noise levels even for a pair of images ($1^{st}$ column). Similar to MLP-based reconstructions~(\Cref{fig:dublinresultsmlp}), recovering facades remains challenging for our models due to occlusions. Although local correlators as MC-CNN~(cosine or +mlp) and NCC ($3^{rd}$ $\&$ $4^{th}$ rows) suffer from noise, they recover facades more faithfully.{The $\frac{B}{H}$ ratios are $\{0.06,0.06,0.12,0.13\}$ for $\{2,3,4,5\}$-views respectively.}}
    
    \label{fig:dublinresults}
\end{figure}

CNNs are known for discarding high frequency signals, a characteristic particularly undesirable  when image content is poor from the start (e.g., historical or satellite imagery). MLP similarity functions retrieve surface granularity and reproduce the objects shapes and their inherent details more faithfully. In Dublin dataset, for instance, the significance of MLP can be clearly pointed to by analyzing the accuracy metrics~(see Table~\ref{tab:statsdublin}). {Moreover, MLP similarities yield the  most complete surface reconstructions for all models especially when the epipolar prior is used~(see \Cref{tab:completeness}). Within this rich context matching scenario (i.e., 4cm GSD in Dublin dataset),~\Ours' per-pixel features are highly discriminative. Hence, even cosine similarities render decent reconstructions~(see \Cref{fig:dublinresults}) and on par with MLPs~(see \Cref{fig:dublinresultsmlp}). On the other hand, evaluating our models against the homographic warping shows that MLP similarities are more sensitive to the matching direction than cosines. The accuracy analysis~(see \Cref{tab:pba:results_all,fig:histogramshomographydublin}) points out a more pronounced performance drop when the MLP is activated ($\simeq 2\times$ w.r.t cosine).} Similarly, in the \textit{out-of-distribution} Montpellier~(see Table~\ref{tab:statsmontpellier}), as long as MLP reigns over $D_{2xGSD}$ and $D_{3xGSD}$ metrics, the cosine remains the more generic similarity measure for $D_{1xGSD}$. This suggests, again, that the MLP generalizes less well than cosine similarity to different sensor characteristics. Last but not least, there is a substantial time gain when building the cost volume with cosine similarity as opposed to feature forwarding to the MLP. 

\subsubsection{The influence of geometric priors}
\begin{figure}[H]
\vspace{-0.6cm}
    \centering
    \includegraphics[height=0.294\textwidth,width=\textwidth]{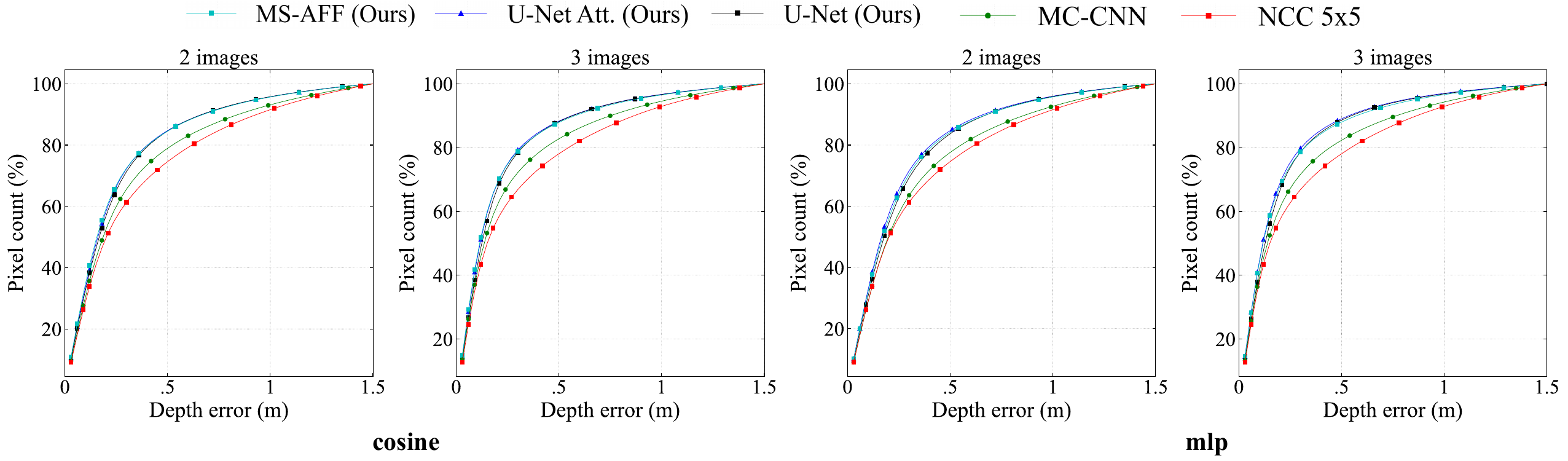}
    \vspace{-0.7cm}
    \caption{\textbf{Homography Geometry Prior vs Cosine/MLP Similarity -- Absolute Error Histogram.} In-distribution Dublin dataset: Histograms corresponding to 2-view and 3-view depth prediction with homographic prior followed by a rotation alignment, with two similarity variants (cosine, MLP). Cosine similarity is slightly more effective than MLP, see Table~\ref{tab:pba:results_all} for explanation. Similarly to the epipolar case, the homographic prior is sensitive to the number of views.  \Ours~are robust and outrun MC-CNN as well as NCC in all scenarios with either geometric priors.}
    \label{fig:histogramshomographydublin}
\end{figure}

\begin{figure}[H]
\vspace{-0.4cm}
    \centering
 \includegraphics[height=0.5\textwidth,width=0.95\textwidth]{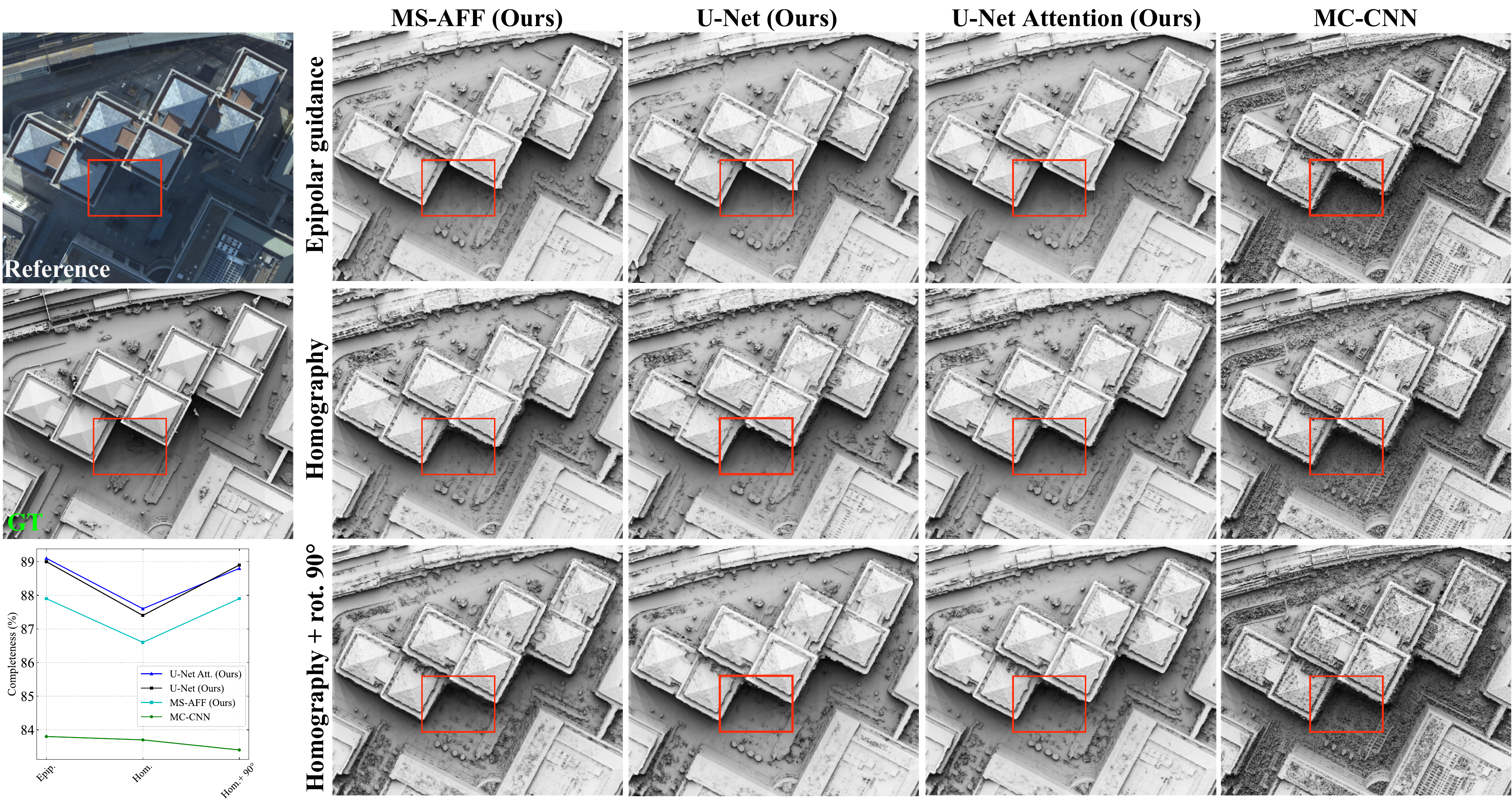}
 \vspace{-0.3cm}
    \caption{\textbf{Directional Sensitivity of Homography Prior -- Visual Assessment.} In-distribution Dublin dataset: Gray-shaded depths predicted from 3 views with epipolar prior, homography prior and homography prior + rotation to roughly align disparities along horizontal lines.  Visual analysis suggests that: (i) epipolar prior is better fit to reconstruct sharp discontinuities, i.e., building edges; and (ii) because applying pure homography induces disparities perpendicularly to the epipolar lines, a direction that the networks have not seen during training using exclusively epipolar training, sharp edges are compromised; (iii) rotating the images by 90$^\circ$ improves the predictions around building edges. The completeness of reconstruction drops for pure homography and mounts for homography with rotation. Quantitative metrics are given in Table~\ref{tab:comaprisonwithhomography}.
    }
    \label{fig:comparison_epipolar_homographic}
\end{figure}

\begin{figure}[htb]
\vspace{-0.6cm}
    \centering
    \includegraphics[width=0.85\textwidth]{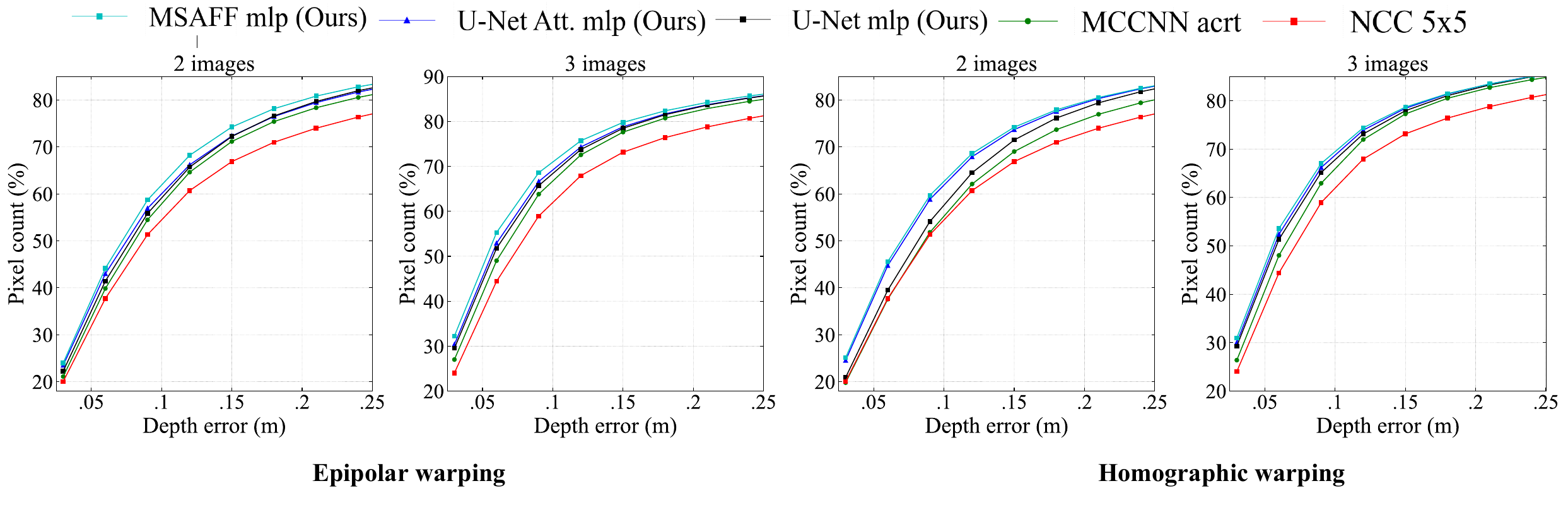}
    \vspace{-0.2cm}
    \caption{\textbf{Epipolar vs Homography Prior -- Absolute Error Histograms.} In-distribution Toulouse dataset: Histograms corresponding to 2- and 3-view depth prediction with geometry priors (epipolar and homographic warping followed by a rotation alignment). MS-AFF variant is most effective across all scenarios. In this experiment, MC-CNN fails to reconstruct lower resolution depths, to obtain reliable depths predictions for high resolutions, we replace it with MS-AFF at resolutions $\frac{\mathcal{R}}{8}$ and $\frac{\mathcal{R}}{4}$ (see Figure \ref{fig:mccnnsfailsatcomplexurbanwithhomography}). 
    }
    \label{fig:histogramstoulouseurban}
\end{figure}

\begin{figure}[H]
    \centering
    \includegraphics[height=1.138\textwidth,width=\textwidth]{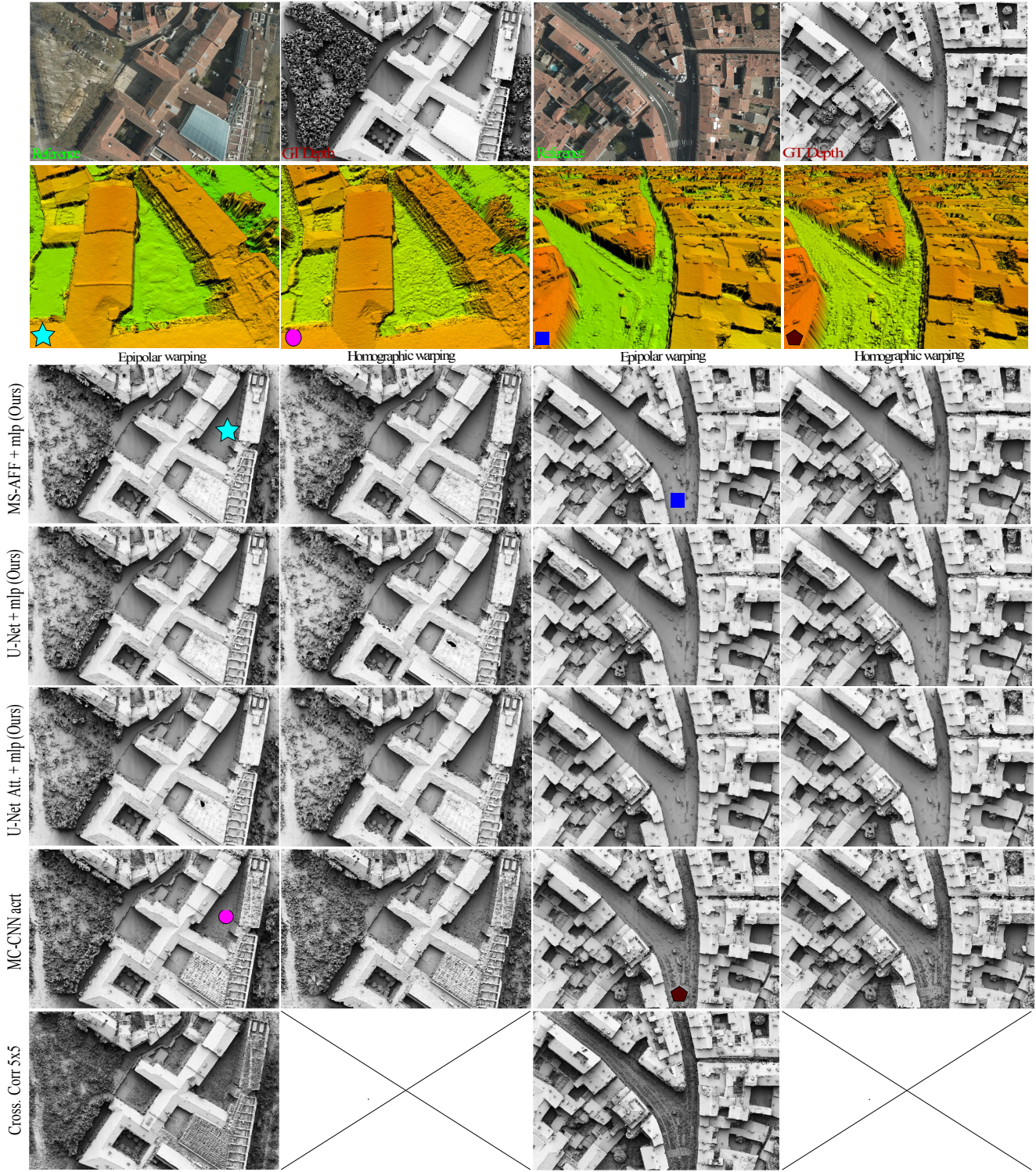}
    \caption{\textbf{Depth Map Visual Quality vs Geometry Priors.} In-distribution Toulouse dataset: Visual comparison between Epipolar and Homographic warping on Toulouse Area (GSD=8cm): Depth maps reconstructed by \Ours~ are noiseless and preserve buildings details compared to MC-CNN and standard correlation. As background can been seen through tree branches, all models being similarity-driven conservatively reconstruct visible surfaces with less ambiguities for \Ours. Flat surfaces are better recovered by our models for both warping scenarios.}
    \label{fig:epipolarhomographictoulouseurban1}
\end{figure}

{In \Cref{tab:pba:results_all} we compare two types of geometric priors: epipolar and homographic. Globally, the epipolar prior performs better than the homographic variant.}
Homographic warping reduces the geometric deformation between a pair of images to an arbitrary translation. However, our models are trained in a specific case where translations occur along horizontal lines (i.e., epipolar geometry).
By incentivizing \OursBase~to globally match large patches of pixels, epipolar-induced pixels arrangements are learnt. This explains why with the homography prior and the disparity following a translation along an arbitrary direction the performance drops (see Figure~\ref{fig:comparison_epipolar_homographic} and Table~\ref{tab:comaprisonwithhomography}). In this scenario, epipolar-based representation learning for multi-view and multi-view stereo is suboptimal.  
To counteract the decline in accuracy, a corrective rotation can be applied. Based on the experiments presented in Figure~\ref{fig:comparison_epipolar_homographic}, we apply the homography prior followed by a $90^{\circ}$ rotation to all of the datasets transforming all of them~(see Figure~\ref{fig:datasetsAndViews}) to a position where disparities are very close to horizontal displacements. The accuracy metrics improve accordingly and approach the accuracy of the epipolar prior in Toulouse (see Table~\ref{tab:pba:toulousestats},~Figures~\ref{fig:histogramstoulouseurban}~\&\ref{fig:epipolarhomographictoulouseurban1}), or even exceed them in Le Mans (see Table~\ref{tab:statslemans}). The exception is Dublin dataset~(see Table~\ref{tab:pba:dublinstats} \& Figure~\ref{fig:histogramshomographydublin}) where there remains an important deviation from the expected displacement along the column coordinate. We deduce that by rotating the Dublin dataset by an adequate angle, similar improvements would be observed. Note that the experiment~(Figure~\ref{fig:comparison_epipolar_homographic},~Table~\ref{tab:comaprisonwithhomography}) does not include the variant with MLP similarities as it severely underperformed compared to the cosine variant. 
 
Shallow networks such as MC-CNN, on the contrary, have small receptive fields and are less affected by disparities in arbitrary directions. Their learning capacity is limited and no geometric epipolar cues are implicitly learnt. Our large receptive field deep CNNs extend learning matching by looking beyond small image {patches} and being perceptive of interactions between pixels in a more global fashion. This boosts matching performance but inevitably pushes these networks to encapsulate geometric priors.  

\begin{table}[H]
    \begin{subtable}[t]{\linewidth}
    \footnotesize
        \centering
         \begin{adjustbox}{max width=\textwidth}
    \begin{tabular}{@{}l@{}>{\raggedright\arraybackslash}lcccccc|cccccc@{}}
    \Xhline{2\arrayrulewidth}
    & \textbf{Configuration} & \multicolumn{6}{c}{\textbf{2 \text{images}}} & \multicolumn{6}{c}{\textbf{3 \text{images}}} \\
    & Architecture & $\mu$$\downarrow$ & $\sigma$$\downarrow$ & $\textit{NMAD}$$\downarrow$ & $D_{1.5\times\textit{GSD}}$$\uparrow$ & $D_{3\times\textit{GSD}}$$\uparrow$ & $D_{5\times\textit{GSD}}$$\uparrow$ & $\mu$$\downarrow$ & $\sigma$$\downarrow$ & $\textit{NMAD}$$\downarrow$ & $D_{1.5\times\textit{GSD}}$$\uparrow$ & $D_{3\times\textit{GSD}}$$\uparrow$ & $D_{5\times\textit{GSD}}$$\uparrow$ \\
    \Xhline{2\arrayrulewidth}
         & MS-AFF \text{cosine} & 0.26 & 0.29 & 0.1 & 21.6\textcolor{red}{\tiny(-1.0)} & 48.7\textcolor{red}{\tiny(-1.2)} & 61\textcolor{red}{\tiny(-1.3)} & 0.22 & 0.28 & 0.08 & 29.2\textcolor{red}{\tiny(-1.1)}  & 59.8\textcolor{red}{\tiny(-0.9)}  & 70.3\textcolor{red}{\tiny(-0.9)}\\ 
         & U-Net \text{cosine}  & 0.27  & 0.29 & 0.11 & 20.2\textcolor{red}{\tiny(-0.9)} &46.0\textcolor{red}{\tiny(-1.2)} & 58.8\textcolor{red}{\tiny(-1.3)} & 0.22 &  0.27 & 0.08 & 26.7\textcolor{red}{\tiny(-1.8)}  & 57.0\textcolor{red}{\tiny(-1.3)}   & 68.8\textcolor{red}{\tiny(-1.1)} \\
         & U-Net Att. \text{cosine}  & 0.26 & 0.29  & 0.1 & 20.8\textcolor{red}{\tiny(-0.9)} & 47.5\textcolor{red}{\tiny(-1.4)} & 60.2\textcolor{red}{\tiny(-1.5)} & 0.21 & 0.27 & 0.08 & 28.5\textcolor{red}{\tiny(-0.8)}  & 59.3\textcolor{red}{\tiny(-0.6)}  & 70.3\textcolor{red}{\tiny(-0.6)} \\
         & MCCNN \text{cosine}~\cite{zbontar2016}  &  0.32 & 0.34 & 0.13 & 18.8\textcolor{green}{\tiny(+0.1)}  & 42.7\textcolor{green}{\tiny(+0.4)} & 54.2\textcolor{green}{\tiny(+0.5)} & 0.27 & 0.32 & 0.1 & 26.2\textcolor{green}{\tiny(+0.5)}  & 53.2\textcolor{green}{\tiny(+0.6)}  & 63.3\textcolor{green}{\tiny(+0.5)}\\
         \hline
         & MS-AFF \text{mlp} & 0.27 & 0.29 & 0.11 & 19.9\textcolor{red}{\tiny(-2.2)} & 45.3\textcolor{red}{\tiny(-4.2)} & 57.7\textcolor{red}{\tiny(-4.4)} & 0.22 & 0.27 & 0.08 & 28.3\textcolor{red}{\tiny(-3.0)} &58.7\textcolor{red}{\tiny(-3.5)} & 69.5\textcolor{red}{\tiny(-3.4)}\\
         & U-Net \text{mlp}  & 0.28 & 0.29 & 0.12 & 19.1\textcolor{red}{\tiny(-2.0)} & 43.7\textcolor{red}{\tiny(-3.5)} & 56.3\textcolor{red}{\tiny(-3.5)}  & 0.22 & 0.26 & 0.08 & 26.4\textcolor{red}{\tiny(-2.8)} & 56.1\textcolor{red}{\tiny(-3.8)} & 68.3\textcolor{red}{\tiny(-3.2)}\\
         & U-Net Att.  \text{mlp}  & 0.26 & 0.29 & 0.11 & 20.1\textcolor{red}{\tiny(-1.5)} & 46.5\textcolor{red}{\tiny(-2.1)} & 59.2\textcolor{red}{\tiny(-2.2)} & 0.21 & 0.26 & 0.08 & 28.4\textcolor{red}{\tiny(-1.9)} &\textbf{59.3}\textcolor{red}{\tiny(-2.2)} & 70.4\textcolor{red}{\tiny(-1.9)} \\ 
         & MCCNN \text{acrt}~\cite{zbontar2016} & 0.33 & 0.34 & 0.14 & 17.8\textcolor{red}{\tiny(-0.8)} & 40.7\textcolor{red}{\tiny(-1.1)} & 52\textcolor{red}{\tiny(-0.9)} & 0.27 & 0.32 & 0.1 & 25.6\textcolor{red}{\tiny(-0.1)} & 52.5\textcolor{green}{\tiny(+0.1)} & 62.6\textcolor{green}{\tiny(+0.1)}\\ 
         \hline
         & NCC $5\times5$~\cite{micmac} & 0.35 & 0.36 & 0.15 & 17.9 & 40.6 & 51.2 & 0.3 &  0.35 &  0.12 & 24.6 & 49.8 & 58.6\\
         \Xhline{2\arrayrulewidth}
    \end{tabular}
    \end{adjustbox}
   \caption{\textbf{Dublin dataset}. Two-view and three-view configurations selected from Figure~\ref{fig:dublinresultsmlp}. Metrics marked in black denote the homography prior results~(see Figure~\ref{fig:histogramshomographydublin} for histograms). For comparison purposes, Colored metrics are differences ($\Delta$) with the epipolar prior metrics (found in Table~\ref{tab:statsdublin}) so that $D_{hom}-D_{epip} = \Delta$. Red means that  epipolar is better than homography and green means the opposite.}
   \label{tab:pba:dublinstats}
    \end{subtable}
    
    \bigskip
    \begin{subtable}[t]{\linewidth}
        \centering
       \begin{adjustbox}{max width=\textwidth}
        \begin{tabular}{@{}l@{}>{\raggedright\arraybackslash}lcccccc|cccccc@{}}
        \Xhline{2\arrayrulewidth}
    & \textbf{Configuration} & \multicolumn{6}{c}{\textbf{2 \text{images}}} & \multicolumn{6}{c}{\textbf{3 \text{images}}} \\
    & Architecture & $\mu$$\downarrow$ & $\sigma$$\downarrow$ & $\textit{NMAD}$$\downarrow$ & $D_{0.75\times\textit{GSD}}$$\uparrow$ & $D_{2\times\textit{GSD}}$$\uparrow$ & $D_{3\times\textit{GSD}}$$\uparrow$ & $\mu$$\downarrow$ & $\sigma$$\downarrow$ & $\textit{NMAD}$$\downarrow$ & $D_{0.75\times\textit{GSD}}$$\uparrow$ & $D_{2\times\textit{GSD}}$$\uparrow$ & $D_{3\times\textit{GSD}}$$\uparrow$ \\
    \Xhline{2\arrayrulewidth}
         & MS-AFF \text{mlp} \textit{\underline{epip}} & 0.16 & 0.25 & 0.05 & 44.2 & 74.2 & 80.8 & 0.14  & 0.24 & 0.03 & 55.3 & 79.8 & 84.2\\
         & U-Net \text{mlp}  \textit{\underline{epip}} & 0.17 & 0.25 & 0.05 & 41.4 & 72.2 & 79.7 & 0.14 & 0.24 & 0.04 & 51.8 & 78.5 & 83.6\\
         & U-Net Att.  \text{mlp}  \textit{\underline{epip}} & 0.17 & 0.26 & 0.05 & 43.0 & 72.3 & 79.4 & 0.14 & 0.24 & 0.04 & 52.9 & 78.8 & 83.7\\
         & MC-CNN \text{acrt}~\cite{zbontar2016}  \textit{\underline{epip}} & 0.18 & 0.26 & 0.05 & 39.8 & 71.1 & 78.3 & 0.15 & 0.24 & 0.04 & 49.0 & 77.6 & 82.9\\
         \hline
         & MS-AFF \text{mlp} \textit{\underline{hom}} & 0.16 & 0.26 & 0.05  & 45.6\textcolor{green}{\tiny(+1.4)} & 74.2\textcolor{red}{\tiny(0.0)} & 80.5\textcolor{red}{\tiny(-0.3)} & 0.14 & 0.25 & 0.04 & 53.6\textcolor{red}{\tiny(-1.7)} & 78.7\textcolor{red}{\tiny(-1.1)} & 83.4\textcolor{red}{\tiny(-0.8)}\\
         & U-Net \text{mlp}  \textit{\underline{hom}} &  0.17 & 0.25 & 0.05 & 39.5\textcolor{red}{\tiny(-1.9)} & 71.5\textcolor{red}{\tiny(-0.7)} & 79.4\textcolor{red}{\tiny(-0.3)} & 0.14 & 0.14 & 0.04 & 51.3\textcolor{red}{\tiny(-0.5)} & 77.9\textcolor{red}{\tiny(-0.6)} & 83.2\textcolor{red}{\tiny(-0.4)}\\
         & U-Net Att.  \text{mlp}  \textit{\underline{hom}} & 0.16 & 0.25 & 0.05 & 44.7\textcolor{green}{\tiny(+1.7)} & 73.7\textcolor{green}{\tiny(+1.4)} & 80.3\textcolor{green}{\tiny(+0.9)} & 0.14 & 0.24 & 0.04 & 52.4\textcolor{red}{\tiny(-0.5)} &78.4\textcolor{red}{\tiny(-0.4)}  &83.4\textcolor{red}{\tiny(-0.3)} \\
         & MC-CNN \text{acrt}~\cite{zbontar2016}  \textit{\underline{hom}} & 0.19 & 0.27 & 0.06 & 37.5\textcolor{red}{\tiny(-2.3)} & 69.0\textcolor{red}{\tiny(-2.1)} & 76.9\textcolor{red}{\tiny(-1.4)} & 0.15 & 0.24 & 0.04 & 48.0\textcolor{red}{\tiny(-1.0)} & 77.2\textcolor{red}{\tiny(-0.4)} & 82.7\textcolor{red}{\tiny(-0.2)}\\ 
         \hline
         & NCC $5\times5$~\cite{micmac} & 0.2 & 0.27 & 0.06 & 37.7 & 66.9 & 74.0  & 0.17 & 0.25  & 0.05 & 44.4 & 73.1 & 78.8\\
         \Xhline{2\arrayrulewidth}
        \end{tabular}
        \end{adjustbox}
     \caption{\textbf{Toulouse dataset}. Epipolar and homography prior 2-view and 3-view configurations selected from Figure~\ref{fig:epipolarhomographictoulouseurban1}. Similar to (a), offsets of homography w.r.t epipolar are colored in red or green. Relevant cumulative histograms are shown in Figure~\ref{fig:histogramstoulouseurban}.}
       \label{tab:pba:toulousestats}
     \end{subtable}  
     \vspace{-0.2cm}
     \caption{\textbf{Epipolar vs Homography Prior -- Accuracy Experiment.} \textit{In-distribution} datasets: \textit{epip} prior consistently outruns the \textit{hom} prior across all multi-view scenarios, except for MC-CNN where the performance of the two is comparable. Additional views are equally beneficial in \textit{epip} and \textit{hom} scenarios. Moreover, cosine similarity is more favorable than MLP when \textit{hom} prior is used, indicating that the MLP trained on epipolar images encodes information specialised to epipolar geometry.}
     \label{tab:pba:results_all}
\end{table} 

\vspace{-0.5cm}
\begin{table}[H]
\centering
\begin{adjustbox}{max width=0.7\textwidth}
\begin{tabular}{@{}l@{\hskip 0.5cm}>{\raggedleft\arraybackslash}ccccccc@{}}
    \Xhline{2\arrayrulewidth}
   Configuration & Model & $\mu$$\downarrow$ & $\sigma$$\downarrow$ & $\textit{NMAD}$$\downarrow$ & $D_{1.5\times\textit{GSD}}$$\uparrow$ & $D_{3\times\textit{GSD}}$$\uparrow$ & $D_{5\times\textit{GSD}}$$\uparrow$ \\
    \hline
    Hom & \multirow{2}{6em}{MS-AFF$_{cos}$} & 0.23 & 0.28 & 0.09 & 26.2 & 55.2 & 66.6\\
    Hom + $90^{\circ}$&   &0.22& 0.28& 0.08 &29.2&59.8&70.3\\\hline
    Hom & \multirow{2}{6em}{U-Net$_{cos}$}  & 0.24 & 0.28 & 0.1 & 24.0 & 51.8 & 63.7\\
    Hom + $90^{\circ}$&  &0.22&0.27& 0.08 &26.7 &57.0 &68.8 \\\hline
    Hom & \multirow{2}{6em}{U-Net Att.$_{cos}$} & 0.24 & 0.28 & 0.09 & 24.5 & 53 & 64.8\\
    Hom + $90^{\circ}$&   & 0.21& 0.27 &0.08& 28.5&59.3&70.3\\\hline
    Hom & \multirow{2}{8em}{MC-CNN$_{cos}$~\cite{zbontar2016}} & 0.27 & 0.32 & 0.1 & 27.2 & 53.7 & 63.2\\
    Hom + $90^{\circ}$&  & 0.27& 0.32& 0.1& 26.2& 53.2& 63.3\\
    \Xhline{2\arrayrulewidth}
\end{tabular}
\end{adjustbox}
  \caption{\textbf{Directional Sensitivity of Homographic Prior.} In-distribution Dublin dataset: Error metrics for \Ours~variants and MC-CNN when homography prior is used. The 90$^\circ$ rotation transforms vertical disparities to approximately horizontal disparities, a setting similar to the one seen by the networks during training. Without the rotation the metrics fall of precision. MC-CNN is least sensitive to direction of disparities because it is a shallow network with limited receptive fields and trained on small image patches. On the contrary, \Ours~implicitly learn epipolar cues making them more vulnerable to generic pixel displacements. Depth map visual assessment is given in Figure~\ref{fig:comparison_epipolar_homographic}.
  }
    \label{tab:comaprisonwithhomography}
\end{table}

\subsubsection{Generalization properties}
\begin{figure}[H]
    \centering
\includegraphics[height=1.037\textwidth,width=\textwidth]{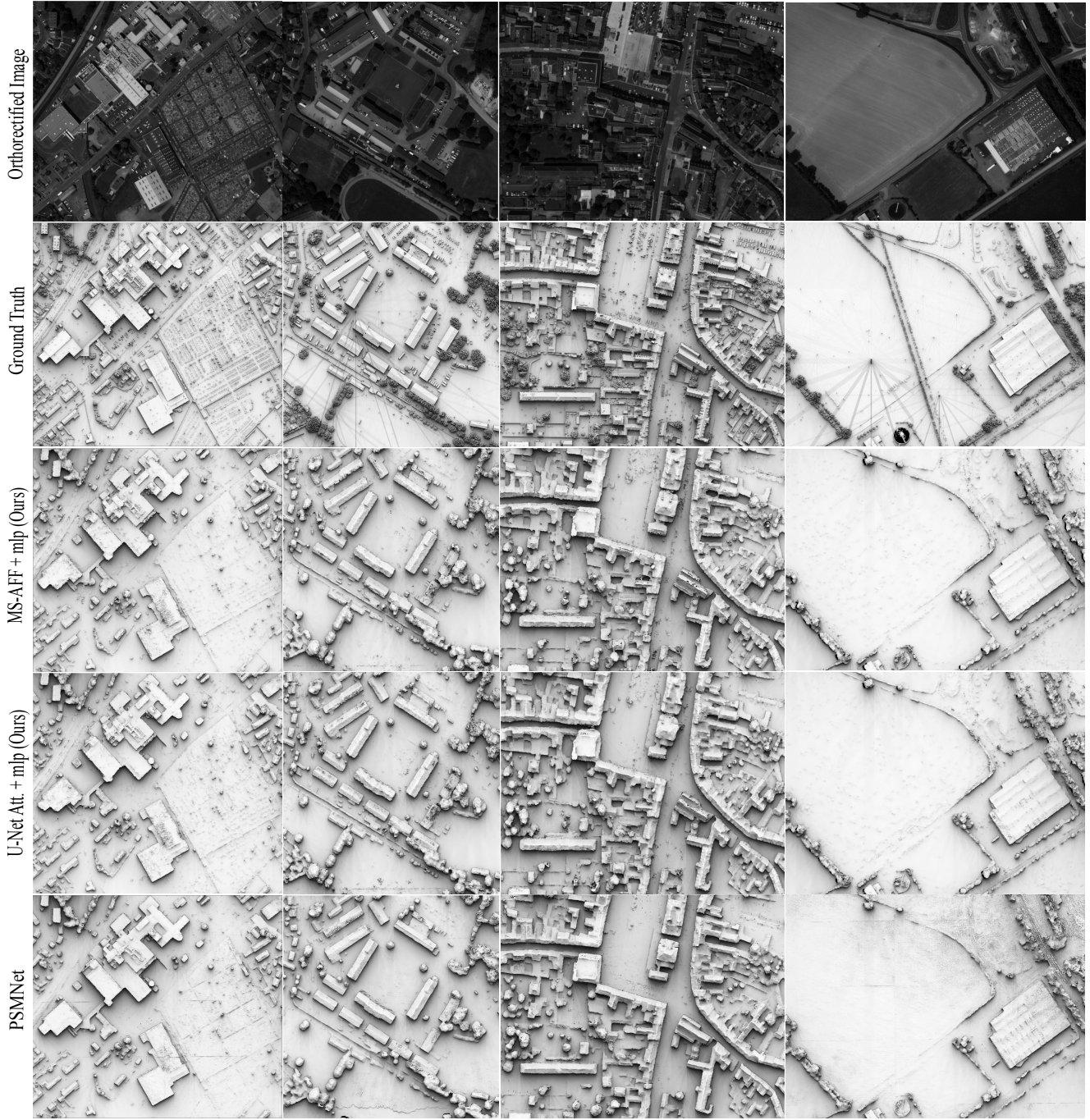}
    \caption{\textbf{Digital Elevation Model Visual Quality vs Generalization.} Out-of-distribution Le Mans Dataset (21~cm): We compare \Ours~with PSMNet~\cite{PSMNet}, RAFT-Stereo~\cite{raftstereo}, MC-CNN acrt\cite{zbontar2016} and standard normalized cross correlation (NCC) \cite{micmac} (see figure~\ref{fig:gsd20cm_mccnandcorrel}). \Ours~yield robust results across different visual contexts (columns) and alleviate noise by grasping similarity cues from larger receptive fields. PSMNet and RAFT-Stereo generate rooftops shapes more faithfully than our models at the cost of dismissing some high frequency details. Note that PSMNet was trained on larger datasets involving several cities and varying buildings shapes. Nonetheless, it introduces some surface artifacts (column 1). \Ours~recover both objects shapes and their inherent details while reducing the noise.}
    \label{fig:gsd20cm}
\end{figure}

\begin{figure}[H]
    \centering
\includegraphics[height=0.622\textwidth,width=\textwidth]{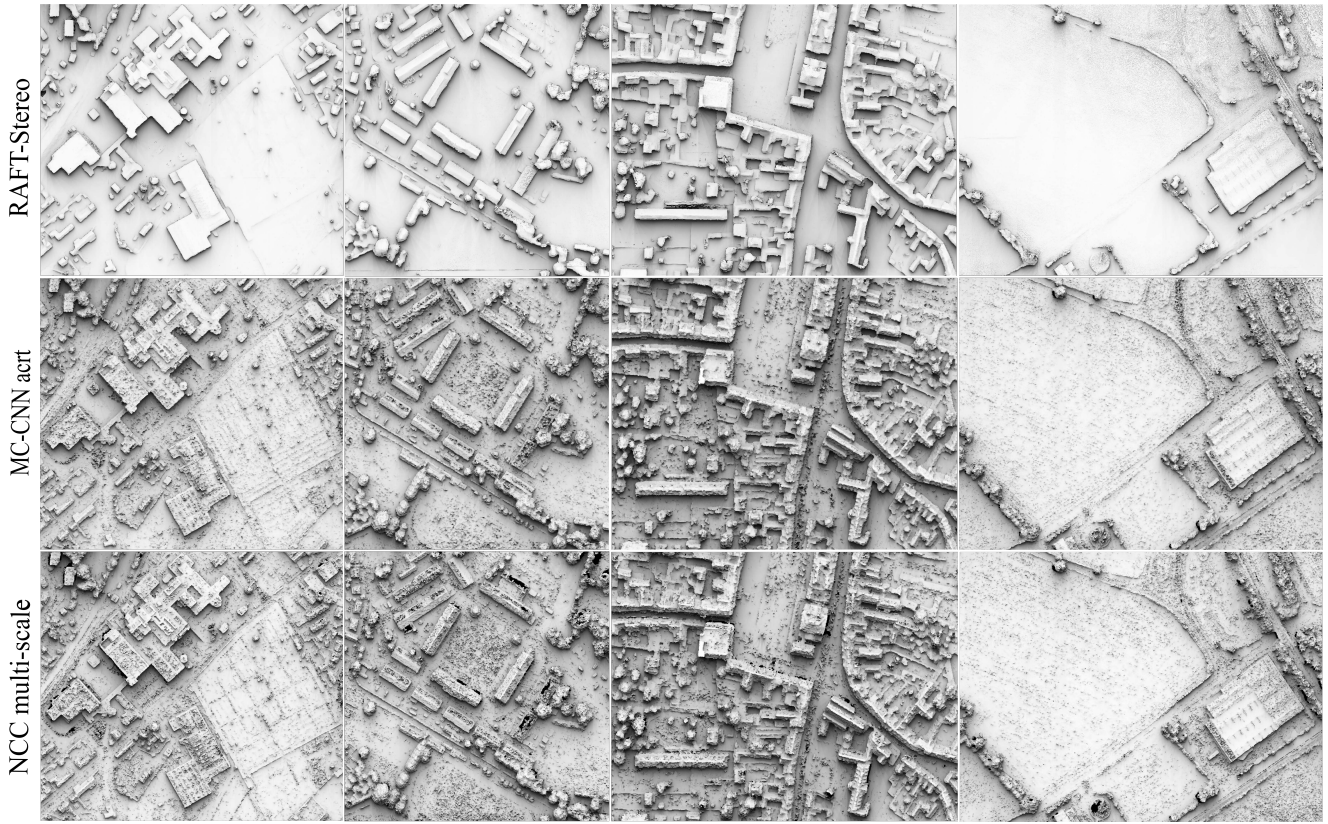}
    \caption{\textbf{Digital Elevation Model Visual Quality vs Generalization.} Out-of-distribution Le Mans Dataset (21~cm): Similarly to PSMNet, RAFT-Stereo yields noise-free reconstructions while ignoring high frequency details. Local correlators reconstructions suffer from noise despite strong regularization. This is due to the limited models' capacity to describe  homogeneous areas. Unlike MC-CNN,~\Ours~overcome homogeneity ambiguities by looking wider into images, facilitating contextual scene understanding.}
    \label{fig:gsd20cm_mccnandcorrel}
\end{figure}

Deep CNN models are sensitive to scale variation unless specifically trained to achieve robustness under scale change. 
MS-AFF embeds multi-scale feature learning explicitly by creating a pyramidal input image and fusing resulting per-scale features via recursive self-attention mechanisms. Consequently,  representations distinctiveness is preserved under a range of GSDs (see \Cref{fig:epipolarhomographictoulouseurban1,fig:gsd20cm,fig:gsd30cm}). Similarly, the U-Net like architectures encode scale variations, but they struggle in shadowed areas (see \Cref{fig:gsd30cm}). The narrow receptive fields of MC-CNN are relatively good at generalization but MC-CNN fails to recover homogeneous buildings rooftops (see \Cref{fig:gsd30cm}) and areas with flat terrain (see\Cref{fig:gsd20cm_mccnandcorrel}), despite heavy cost volume regularisation. MS-AFF is the best performing model in terms of generalization. It combines robustness, surface reconstruction faithfulness and high accuracy on \textit{out-of-distribution} datasets.

Dense image matching pipelines that learn shape cues implicitly, fit the features to the specific geometries of the training dataset \cite{WU2024103715,Chebbi_2023_CVPR}. Hence, deep regression models like PSMNet~\cite{PSMNet} are excellent interpolators when exposed to familiar scenes with similar objects shapes and sufficiently rich geometrical patterns. This over-specialization undermines their robustness under unseen GSD-size acquisitions. Note the performance drop that PSMNet encounters on Montpellier dataset (see Table~\ref{tab:statsmontpellier} and Figure~\ref{fig:gsd30cm}). {This observation does not hold for RAFTStereo~\cite{raftstereo}, which demonstrates strong generalization capabilities at this GSD level. We attribute this performance stability to the large amount of datasets used during training}. \Ours~ are robust under the same GSD decrease and leverage powerful similarity cues that transfer well through varying landscapes, radiometries and contexts.

\begin{table}[H]
    \centering
    \begin{tabular}{@{}l@{}>{\raggedright\arraybackslash}lcccccc@{}}
    \Xhline{2\arrayrulewidth}
    & \textbf{Architecture} & \textbf{$\mu$$\downarrow$} & $\sigma$$\downarrow$ & $\textit{NMAD}$$\downarrow$ & $D_{1\times\textit{GSD}}$$\uparrow$ & $D_{2\times\textit{GSD}}$$\uparrow$ & $D_{3\times\textit{GSD}}$$\uparrow$\\
    \Xhline{2\arrayrulewidth}
     & MS-AFF \text{cos} \textit{epip} & 0.44 & 0.46 & 0.19 & 55.8 & 75.8 & 85.4\\
     & MS-AFF \text{mlp} \textit{epip} & 0.44 & 0.45 & 0.19 & 55.1 & 76.0 & 85.9\\
     \hline
     & U-Net \text{cos}  \textit{epip} & 0.46  & 0.47 & 0.2 & 54.5 & 74.9 & 84.6  \\
     & U-Net \text{mlp}  \textit{epip} & 0.47 & 0.46 & 0.21 & 51.7 & 73.8 & 84.6\\
     \hline
     & U-Net Att.  \text{cos}  \textit{epip} & 0.49 & 0.49  & 0.22 & 51.7 & 72.5 & 82.7\\ 
     & U-Net Att.  \text{mlp}  \textit{epip} & 0.44 & 0.45  & 0.21 & 51.5 & 73.3 & 83.7\\ 
     \hline
     & MC-CNN \text{cos}~\cite{zbontar2016}  \textit{epip} & 0.51 & 0.46 & 0.23 & 46.0 &  71.6 & 83.6 \\ 
     & MC-CNN \text{mlp}~\cite{zbontar2016}  \textit{epip} & 0.44 & 0.44 & 0.19 & 55.3 & 77.0 & 86.7 \\ 
     \hline
     & NCC $5\times5$~\cite{micmac} \textit{multi-view} & 0.43 & 0.43 & 0.19 & 53.8 & 77.4 & 87.6\\
     & PSMNet~\cite{PSMNet} & 0.60 & 0.49 & 0.30 & 36.1 & 62.2 & 78.0 \\
     & RAFTStereo~\cite{raftstereo} & 0.43 & 0.45 & 0.18 & 58.4 & 77.4 & 86.3 \\
    \Xhline{2\arrayrulewidth}
    \end{tabular}
    \caption{\textbf{Generalization -- Three-View Experiment.} Out-of-distribution Montpellier Tri-Stereo Dataset (GSD=32~cm). All likelihood-based models including standard correlation outperform PSMNet. RAFT-Stereo ranks first and renders accurate reconstructions. MS-AFF \textit{cosine} ranks second on $D_{1\times\textit{GSD}}$. Multi-view NCC $5\times5$ yields the best results starting from $D_{2\times\textit{GSD}}$. Note that images and Lidar acquisitions are asynchronous. Persistent objects may be comparable between the two modalities. As MC-CNN~ (Figure~\ref{fig:gsd30cm}) and NCC~(Figure \ref{fig:groundtruth_issue}) are prone to noise, there is no visual structural similarity between their relative surfaces and the ground truth LIDAR DEM. Small objects that~\Ours~neglects unfavorably contribute to error histograms and are considered close to ground truth. }
    \label{tab:statsmontpellier}
\end{table}

\section{Conclusions}

In this study, we introduce~\Ours, a generalized multi-view dense image matching version of \OursBase, which outperforms state-of-the-art similarity matching networks on all examined metrics and end-to-end methods in terms of generalization. Our approach addresses the multi-view matching problem in a comprehensive fashion. First, it takes advantage of epipolar geometry simplicity to learn powerful, context-aware and locality-free similarity cues. Then, it exploits geometry priors to harness these properties by solving for the rotational transform between a pair of images either by epipolar or homographic warping. By doing so, we enable deep multi-view similarity matching without explicitly retraining our networks in multi-view. The introduction of our lightweight MS-AFF model, which incorporates explicit multi-scale feature learning, results in robust embeddings that exhibit remarkable generalizability across diverse and previously unseen satellite imagery and landscapes.  While homographic warping demonstrates considerable potential, we advocate for further research to refine epipolar-aware embeddings for effectively addressing generic translational pixel displacements.

\begin{figure}[H]
    \centering
\includegraphics[trim={0 128cm 0 0},clip, width=1.0\textwidth]{figures/ZOOM_GSD_30CM_MPL_ADDITIONAL_COMPRESS.pdf}
    \phantomcaption
\end{figure}

\begin{figure}[H]
\ContinuedFloat
    \centering
\includegraphics[trim={0 0 0 128cm},clip,width=1.\textwidth]{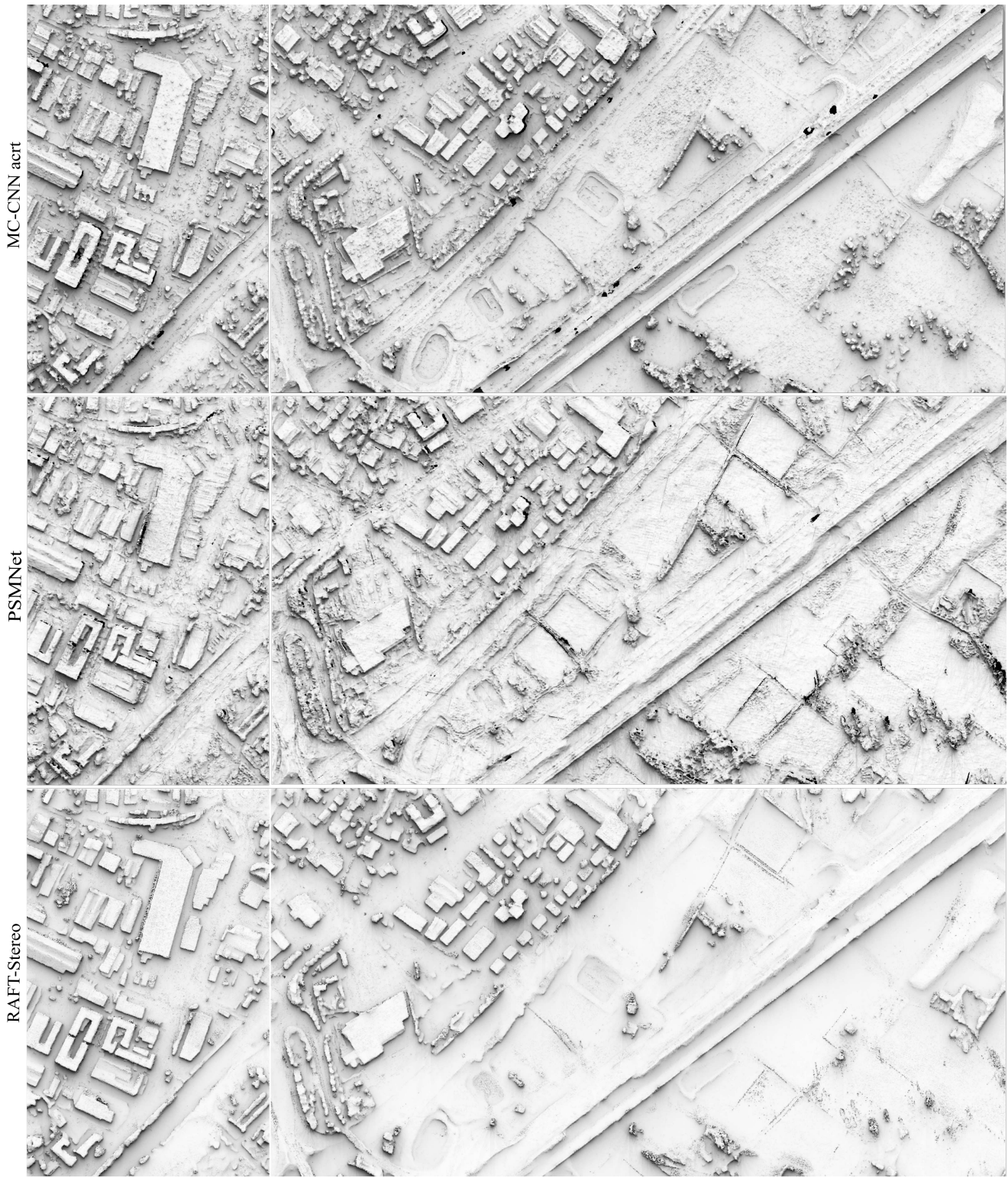}
    \caption{\textbf{Digital Elevation Model Visual Quality vs Generalization.} Out-of-distribution Montpellier Tri-Stereo Dataset (32~cm): \Ours~show decent generalization properties on \textbf{unseen} high resolution satellite tri-stereo images. MS-AFF benefits its explicit multiscale architecture~\cite{Chebbi_2023_CVPR} to learn scale aware fused embeddings that are sufficiently robust to maintain matching performance under poor information content. U-Net (not shown here) and U-Net attention follow the same trend but are sensitive to shadowed regions resulting in unmatched areas (black spots). As MC-CNN is incapable of learning dataset specific contexts, it seems to yield consistent reconstructions. PSMNet drops in performance with quasi-absent buildings details and heavily smoothed objects outlines. RAFT-Stereo separates well buildings footprints and renders noiseless reconstructions though missing some details. Note the Lidar ground truth presents inaccuracies due to an important temporal difference between the image and Lidar acquisitions.}
    \label{fig:gsd30cm}
\end{figure}

\newpage
\bibliographystyle{elsarticle-num}
\bibliography{ISPRS}







\end{document}